\documentclass{article}

\PassOptionsToPackage{numbers, square}{natbib}

\usepackage[final]{neurips_2022}

\usepackage{caption}
\usepackage{subcaption}
\usepackage{amsmath}
\usepackage{soul}
\usepackage{enumitem}
\usepackage{float}
\usepackage{wrapfig}
\usepackage{graphicx}
\usepackage{wrapfig}

\usepackage[utf8]{inputenc} 
\usepackage[T1]{fontenc}    
\usepackage{hyperref}       
\usepackage{url}            
\usepackage{booktabs}       
\usepackage{amsfonts}       
\usepackage{nicefrac}       
\usepackage{microtype}      
\usepackage{xcolor}         

\usepackage{amsmath,amsfonts,bm}

\newcommand{\reddum}[1]{\textcolor{black}{#1}}
\newcommand{\bluedum}[1]{\textcolor{black}{#1}}

\def\secref#1{section~\reddum{\ref{#1}}}

\def\eqref#1{equation~\bluedum{\ref{#1}}}

\def\1{\bm{1}}

\DeclareMathAlphabet{\mathsfit}{\encodingdefault}{\sfdefault}{m}{sl}
\SetMathAlphabet{\mathsfit}{bold}{\encodingdefault}{\sfdefault}{bx}{n}

\usepackage{amsmath}
\usepackage{amssymb}
\usepackage{xcolor}

\usepackage{siunitx}
\usepackage{textcomp}

\usepackage{todonotes}

\newcommand{\SKIP}[1]{}
\newcommand{\real}{\mathbb{R}}
\newcommand{\dataset}{\mathcal{D}}
\newcommand{\pointset}{\mathcal{P}}
\newcommand{\ictset}{\mathcal{K}}
\newcommand{\param}{\theta}

\newcommand{\ipspace}{\mathcal{X}}

\newcommand{\opspace}{\mathcal{Y}}
\newcommand{\suploss}{\ell_{\rm s}}
\newcommand{\usuploss}{\ell_{\rm us}}
\newcommand{\totalloss}{\ell}
\newcommand{\mixop}{\texttt{Mix}_{\pmb{\lambda}}}

\newcommand{\linmsr}{\texttt{NonLin}}
\newcommand{\ictx}{\texttt{ICT}}

\newcommand{\mixup}{{\tt mixup}}

\usepackage{xspace}
\makeatletter
\DeclareRobustCommand\onedot{\futurelet\@let@token\@onedot}
\def\@onedot{\ifx\@let@token.\else.\null\fi\xspace}
 
\def\ie{\emph{i.e}\onedot}

\makeatother

\newcommand{\negspace}[1]{\vspace{#1}}

\title{When does mixup promote local linearity in learned representations?}

\author{%
  Arslan Chaudhry \thanks{The work was done when the author was with Google Research. The author is now at DeepMind.} \\
  DeepMind \\
  \texttt{arslanch@google.com} \\
  \And
  Aditya Krishna Menon \\
  Google Research \\
  \texttt{adityakmenon@google.com} \\
  \And
  Andreas Veit \\
  Google Research \\
  \texttt{aveit@google.com} \\
  \And
  Sadeep Jayasumana \\
  Google Research \\
  \texttt{sadeep@google.com} \\
  \And
  Srikumar Ramalingam \\
  Google Research \\
  \texttt{rsrikumar@google.com} \\
  \And
  Sanjiv Kumar \\
  Google Research \\
  \texttt{sanjivk@google.com} \\
}

\begin{document}

\maketitle

\begin{abstract}
\mixup{} is a regularization technique that artificially produces new samples using convex combinations of original training points. This simple technique has shown strong empirical performance, and has been heavily used as part of semi-supervised learning techniques such as mixmatch~\citep{berthelot2019mixmatch} and interpolation consistent training (ICT)~\citep{verma2019interpolation}. In this paper, we look at \mixup{} through a \emph{representation learning} lens in a semi-supervised learning setup. 
In particular, we study the role of \mixup{} in promoting linearity in the learned network representations. Towards this, we study two questions: 
(1) how does the \mixup{} loss that enforces linearity in the \emph{last} network layer propagate the linearity to the \emph{earlier} layers?; 
and
(2) how does the enforcement of stronger \mixup{} loss on more than two data points affect the convergence of training? 
We empirically investigate these properties of \mixup{} on vision datasets such as CIFAR-10, CIFAR-100 and SVHN.
Our results show that supervised \mixup{} training does not make \emph{all} the network layers linear;
in fact the \emph{intermediate layers} become more non-linear during \mixup{} training compared to a network that is trained \emph{without} \mixup{}. 
However, when \mixup{} is used as an unsupervised loss, we observe that all the network layers become more linear resulting in faster training convergence. 
\end{abstract}

\section{Introduction} \label{sec:intro}
While models learned via empirical risk minimization (ERM) ~\citep{vapnik1998} tend to perform well on test data that are similar to training data, predictions can change significantly when the samples are chosen outside the training distribution. For improved generalization, typically \emph{data augmentation} techniques are used to generate new training examples near the neighborhood of the original training samples through simple transformations~\citep{Simard1996TransformationII}. 
Such techniques play a critical role in training deep neural networks that have shown great success~\citep{Krizhevsky2012,Szegedy2015,He2016DeepRL,Devlin2018}.
While popular data augmentation techniques such as filtering and cropping on images tend to produce samples near the vicinity of training samples, they are domain-specific and require expert knowledge in generating augmentations. 

On the contrary, \emph{\mixup{}}~\citep{Zhang2017} -- an augmentation technique that generates new training data by linearly interpolating the original training points -- is applicable in various domains without requiring expert domain knowledge. \mixup{} has shown strong performance in both supervised~\citep{Zhang2017} and semi-supervised~\citep{verma2019interpolation, berthelot2019mixmatch} learning setups by allowing the model to learn \emph{better network representations}. In the context of supervised learning, the representations learned through \mixup{} regularization are shown to improve generalization of the network and also its robustness to corrupt labels~\citep{Zhang2017}. The general idea of linearly combining the two vectors in \mixup{} can then be extended to the intermediate layers of the network resulting in even better representations~\citep{verma2019manifold}. Additionally, \mixup{} training has been shown to improve network calibration for both in- and out-of-distribution data~\citep{thulasidasan2020mixup}. 

In the context of semi-supervised learning, \mixup{} regularization, also known as \emph{interpolation consistency training} (\emph{ICT}), is shown to improve the performance of a learner significantly. Specifically, \citet{verma2019interpolation}, showed that \mixup{} training encourages consistency regularization which is an effective unsupervised learning signal explored in many works~\citep{sajjadi2016regularization,laine2016temporal,tarvainen2017mean,miyato2018virtual}. Other related methods include mixmatch~\cite{berthelot2019mixmatch} that rely on mixing labeled and unlabeled samples, and methods relying on pseudo-labels~\cite{sohn2020fixmatch} and data augmentation~\cite{xie2020unsupervised}.

While \mixup{} regularization has brought performance benefits across the board, it is less clear how 
\emph{network representations} differ between
a network trained with and without \mixup{}. 
Some theoretical works have studied the improved generalization of \mixup{} through the lens of regularization effects it brings to the network training. \citet{carratino2020mixup} argued that \mixup{} regularizes the Jacobian of the network resulting in a function with a low Lipschitz constant. Similarly, \citet{zhang2020does} showed that, in a two-layer ReLU network, \mixup{} training reduces the complexity of the hypothesis class leading to better generalization. While these works point to some regularization effects of \mixup{}, the training dynamics of a network trained under \mixup{} regularization are not fully clear.

Our work is an empirical study that aims to provide some clues as to how network representations evolve during \mixup{} training. Specifically, $1$) we show that \mixup{} tends to make the first and the last layer of a network more linear, but does so at the expense of making the intermediate layers more non-linear, compared to a network that is trained without \mixup{}. $2$) We show that when \mixup{} is used as an unsupervised loss, ICT, all layers tend to become more linear. Finally, $3)$ we show that enforcing a stronger linearity in \mixup{}, by means of using more than two mixing points, leads to more linear representations that manifest in faster convergence to a specified test accuracy with less labeled examples in a semi-supervised setup. 

\section{Setup} \label{sec:setup}
\negspace{-2mm}
Let $\ipspace \in \real^D$ and $\opspace \in \real^C$ be input and output spaces, respectively. 
We adopt a standard semi-supervised learning setup where we assume access to a small labeled dataset $\dataset_s=\{(x_i, y_i)\}_{i=1}^{n_s}$, and a large unlabeled dataset $\dataset_{\rm us}=\{u_i\}_{i=1}^{n_{\rm us}}$, with $n_{s} \ll n_{\rm us}$ are the number of examples in the labeled and unlabeled datasets, respectively.
%
The training objective is to learn a model $f_{\param}: \ipspace \mapsto \opspace$, parameterized by $\param \in \real^P$ (a neural network in our case), that performs well on a held out test set $\dataset_{\rm te}$ by minimizing the following loss:
$$\totalloss = \suploss + w_t \cdot \usuploss,$$
where $\suploss: \opspace \times \opspace \mapsto \real_{\geq 0}$ is a supervised loss computed on $\dataset_s$ (typically a cross-entropy loss for classification tasks), 
$\usuploss$ is an unsupervised loss defined on $\dataset_{\rm us}$, and 
$w(t)$ is the weight of unsupervised loss at the $t$-th iteration of stochastic gradient descent (SGD).
We define the {\em mixup operation} as a convex operation parameterized by $ \pmb{\lambda} \in \real_{\geq 0}^{K}$, on a set $\pointset = \{p_1, \cdots, p_K\}$ as:
\negspace{-2mm}
\begin{equation} \label{eq:mixup_op}
    \mixop(\pointset) = \sum_{i=1}^K \lambda_i p_i, 
\end{equation}

such that $\sum_{i=1}^K \lambda_i = 1$, and $\lambda_i \geq 0$ for all $i \in \{1, \cdots, K\}$.
A sample from a $K$-th order Dirichlet distribution with parameters $\alpha_1, \cdots, \alpha_K > 0$, defines a valid $\pmb{\lambda}$ for the mixup operation.
When the mixup operation is used on the supervised loss $\suploss$, the examples $x \in \ipspace$ and targets $y \in \opspace$ are modified as: 
\begin{align*}
    x_{\rm m} &= \mixop(x_1, \cdots, x_K), \\
    y_{\rm m} &= \mixop(y_1, \cdots, y_K),    
\end{align*}
at each training step of the SGD. 
For the case of $K=2$, the supervised loss becomes the standard \mixup{} training as proposed by \citet{Zhang2017}.
Similarly, we can also use the mixup operation to define an unsupervised loss on the unlabeled examples. Concretely, given $K$ unlabeled examples $\mathcal{K} = \{u_1, \cdots, u_K\}$, the unsupervised loss $\usuploss$ is defined as,
\negspace{-2mm}
\begin{equation} \label{eq:ictx}
    \usuploss = D\biggr( f\bigl(\mixop(\ictset)\bigr) - \mixop\bigl(f(u_1), \cdots, f(u_K)\bigr) \biggr), 
\end{equation}

where $D$ is some distance metric; $L_2$ distance is used in this work. Intuitively, the unsupervised loss enforces consistent predictions on the interpolated unlabeled examples which \citet{verma2019interpolation} found to be a good learning signal for \emph{consistency regularization} in semi-supervised setups. 

\section{Experiments} \label{sec:experiments}
\negspace{-2mm}
\paragraph{Setup} We assume standard semi-supervised learning datasets (CIFAR-10, CIFAR-100, and SVHN) and experiment with different labeled dataset sizes. For CIFAR-10 and SVHN, we consider labeled datasets of sizes $\{250, 1000, 2000, 40000\}$. For CIFAR-100, we only consider a single labeled dataset of size $10$K. 
We use a `Wide ResNet-28' architecture similar to \citep{berthelot2019mixmatch}.
We denote the output of the initial convolution layer as `Layer 0', and the outputs of the next three Wide ResNet blocks are referred to as `Layer 1-3', followed by the output of average pooling layer that is denoted by `Layer~4`. Finally the logits layer is referred to as `Layer 5'.
Under this definition of a `layer', we monitor the representation after each layer. 
The batch size is set to $128$ and the Adam optimizer~\citep{kingma2014adam} is used with learning rate = $0.002$, $\beta_1$ = 0.9 and $\beta_2$ = 0.999. 
The weight ($w_t$) of the unsupervised loss follows a schedule such that its value linearly increases from $0$ to a maximum of $100$ in $40$\% of training iterations that are kept fixed for all datasets and for all dataset sizes. 
The linearity of network `layers' is periodically reported throughout the network training on a held out test set.
The $\alpha$ for the Dirichlet distribution is set to $0.5$ for CIFAR-10 and CIFAR-100, and $0.1$ for SVHN.
Each method is run with $5$ different random seeds and the performance is averaged across these runs, and error bars are reported for convergence plots. 

\paragraph{Baselines} We compare $3$ different baselines. \textbf{ERM} refers to standard empirical risk minimization \emph{without} \mixup{}. \textbf{\mixup{}} refers to training using two \mixup{} points. In both these setups, only the supervised labeled set $\dataset_{\rm s}$ is used and $\dataset_{\rm us}$ is not used. \textbf{ICT[K]} refers to unsupervised \mixup{}, as defined in \eqref{eq:ictx}, with $K \in \{2, 3, 4\}$ referring to the number of examples mixed together. For this setup, the ICT loss is computed on the $\dataset_{\rm us}$ and standard cross-entropy loss computed on $\dataset_s$.

\subsection{Non-Linearity in Network Representations} \label{sec:linearity_prop}
\negspace{-2mm}
We first investigate the effect of \mixup{} on the network representations at different layers during training. Specifically, we take a look at the effect of \mixup{} training on the (non-)~linearity of network representations. We specifically chose to study the linearity of representations because the \mixup{} theory~\citep{carratino2020mixup,zhang2020does} suggests that \mixup{} trained networks tend to be more linear than their non-\mixup{} counterparts. 
We define the non-linearity in a layer $l$ at the $t$-th training step by:
\negspace{-2mm}
\begin{equation} \label{eq:linearity}
\begin{split}
    \linmsr_t(l) = \sum_{(x_1, x_2) \sim \dataset_{te}} D\biggl(&\hat{f}^{l}_t(\mixop(x_1, x_2)) - \mixop(\hat{f}^{l}_t(x_1), \hat{f}^{l}_t(x_2)) \biggr), 
\end{split}
\end{equation}
where $\dataset_{\rm te}$ is a held out test set, and $\hat{f}^l_t(\cdot)$ is the normalized output of layer $l$ at the $t$-th training step. 
We normalize the output of a layer to have a unit norm, \ie $\hat{f}^l_t= \frac{f^l_t}{\|f^l_t\|}$, to allow for a fair comparison of the non-linearity across different layers.
\emph{Lower values imply a more linear layer}. 

\begin{figure*}[t]
\def \LABELS {250}
\def \CTWEIGHT {100}	
\centering
\begin{subfigure}{.33\textwidth}
      \centering
      \includegraphics[width=.99\linewidth]{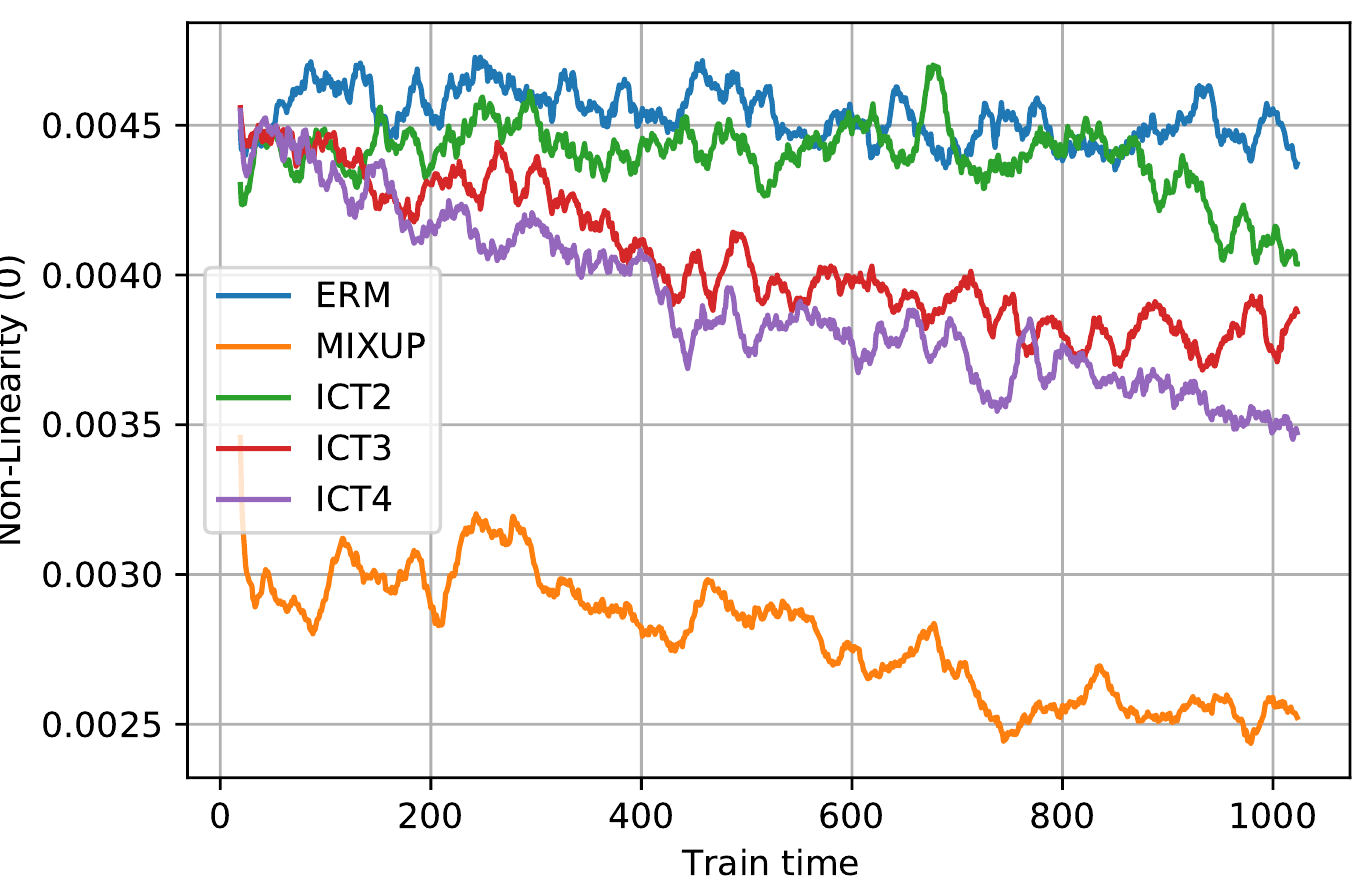}
      \caption{\small Layer 0}
\end{subfigure}\hfill
\begin{subfigure}{.33\textwidth}
      \centering
      \includegraphics[width=.99\linewidth]{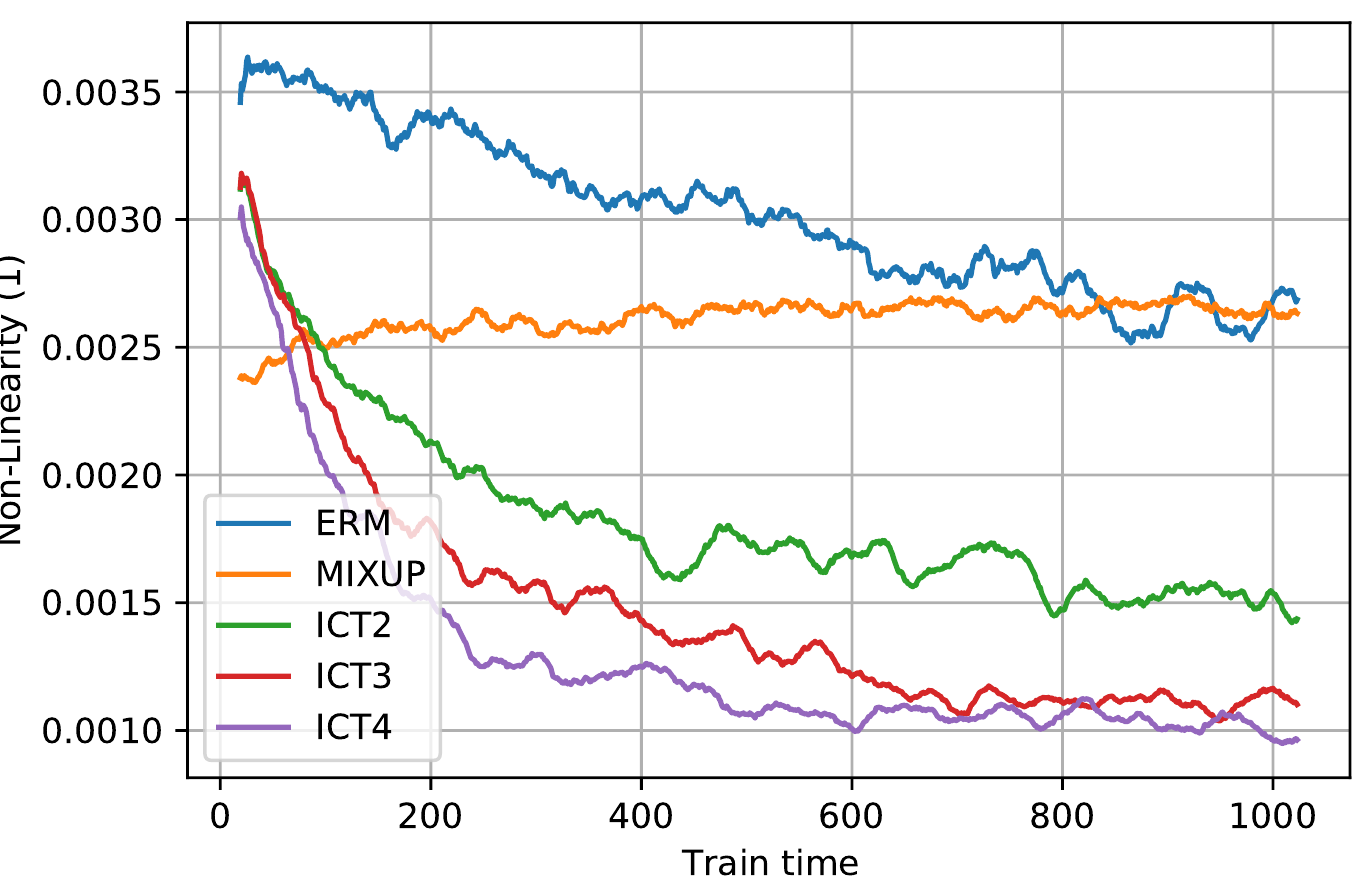}
      \caption{\small Layer 1}
\end{subfigure}\hfill
\begin{subfigure}{.33\textwidth}
      \centering
      \includegraphics[width=.99\linewidth]{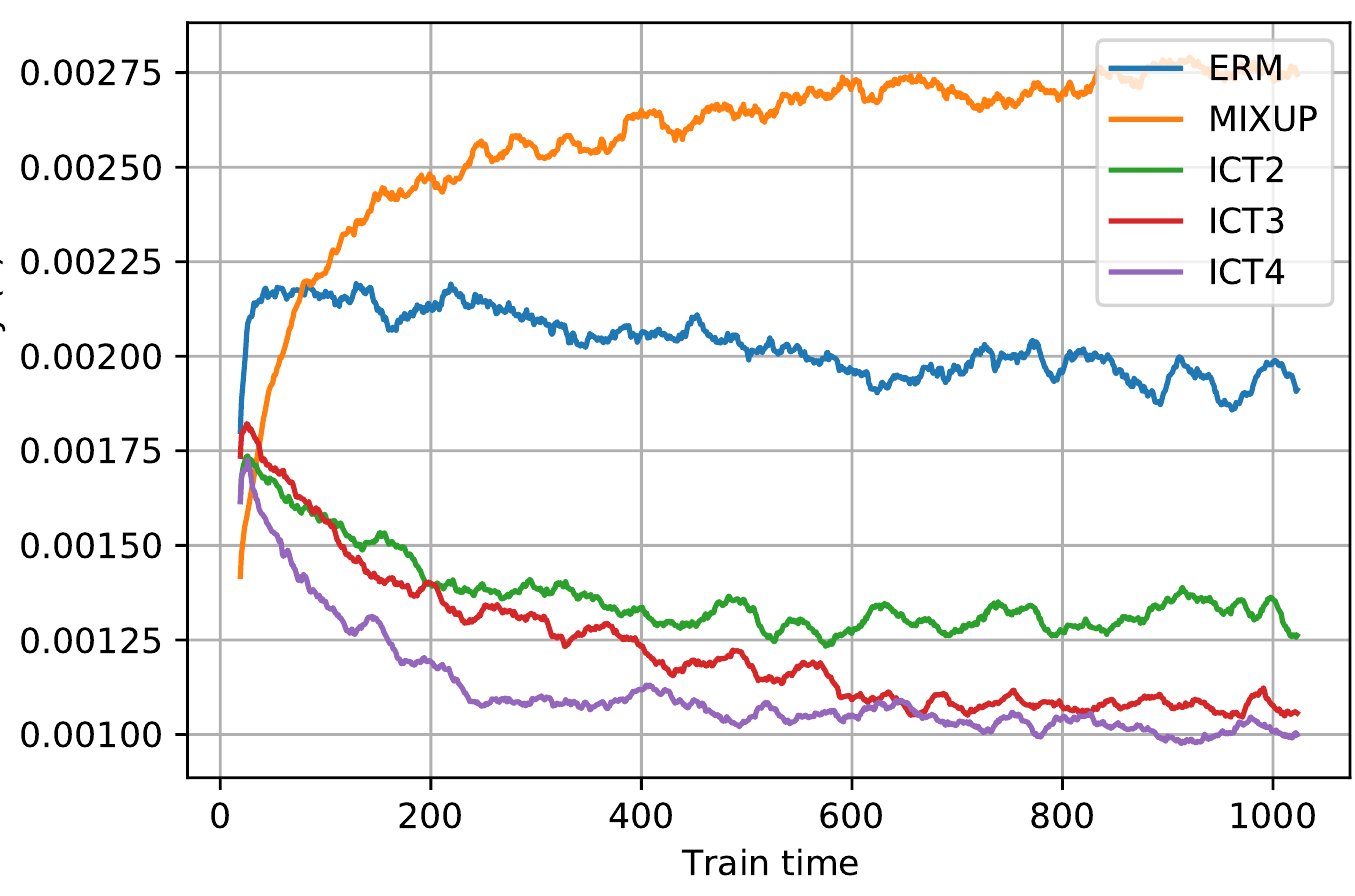}
      \caption{\small Layer 2}
\end{subfigure}

\begin{subfigure}{.33\textwidth}
      \centering
      \includegraphics[width=.99\linewidth]{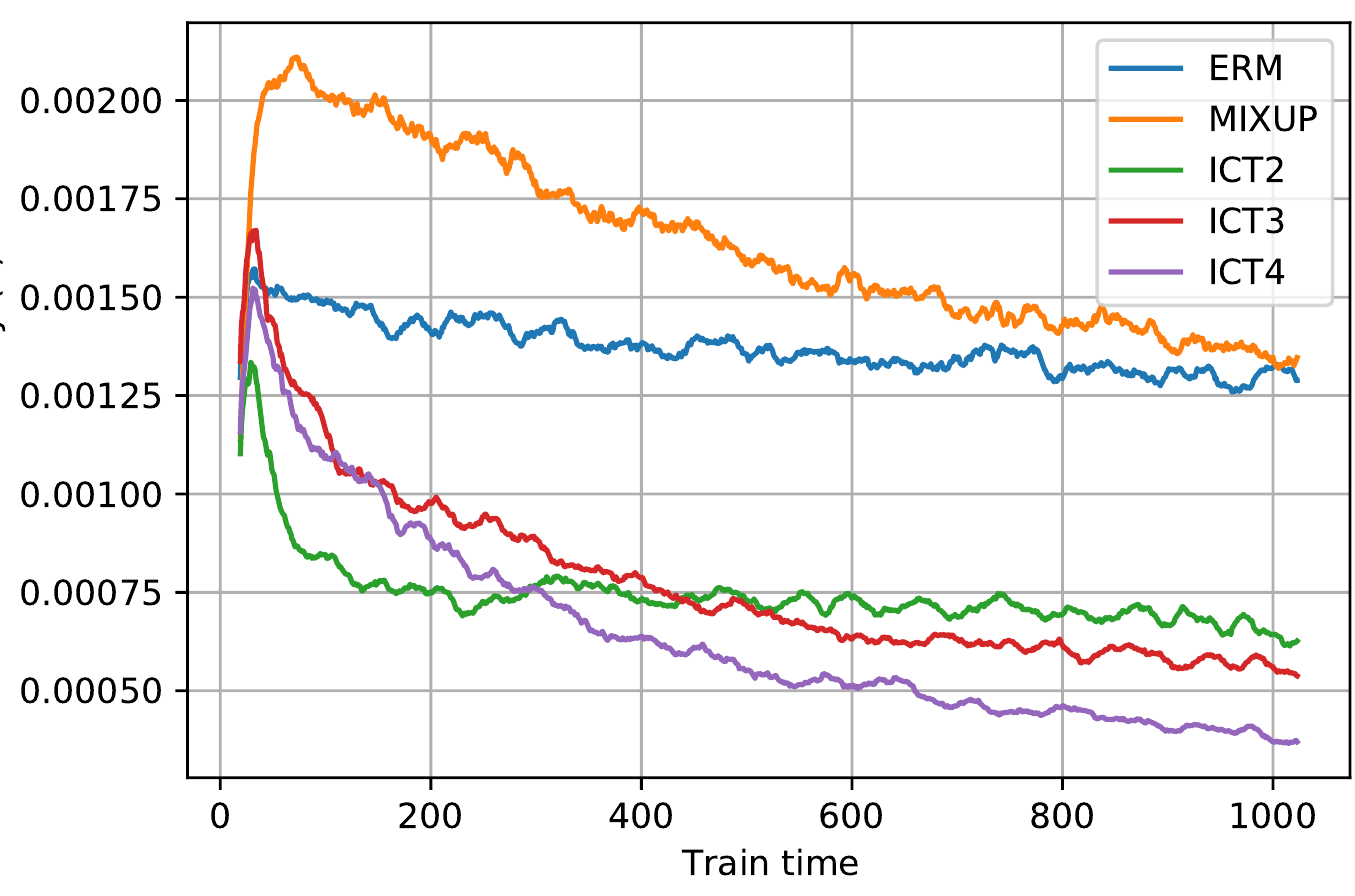}
      \caption{\small Layer 3}
\end{subfigure}\hfill
\begin{subfigure}{.33\textwidth}
      \centering
      \includegraphics[width=.99\linewidth]{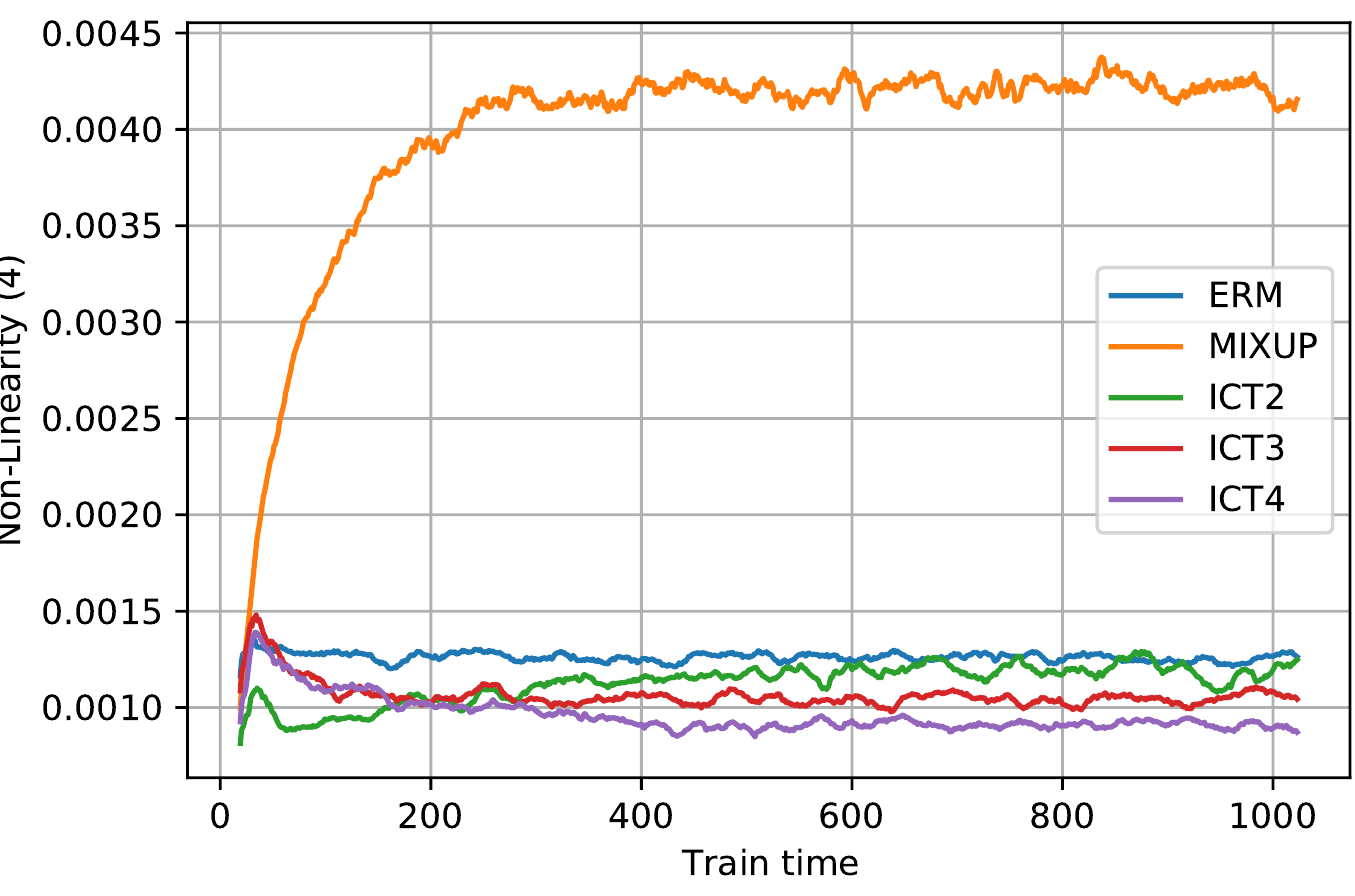}
      \caption{\small Layer 4}
\end{subfigure}\hfill
\begin{subfigure}{.33\textwidth}
      \centering
      \includegraphics[width=.99\linewidth]{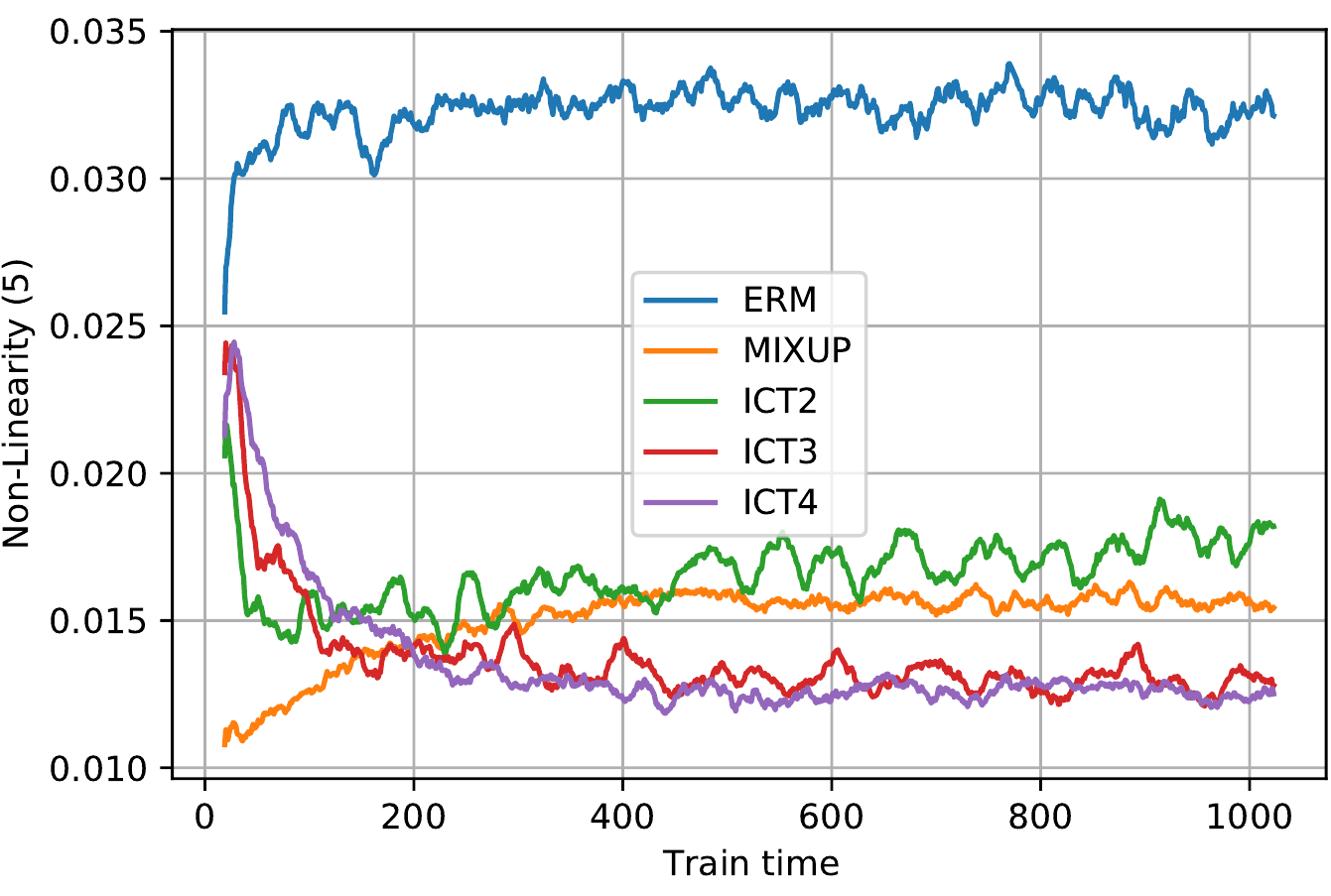}
      \caption{\small Layer 5}
\end{subfigure}

\caption{\textbf{CIFAR-10}: Linearity propagation in different layers with \LABELS\ labeled examples and consistency weight of \CTWEIGHT. ERM and \mixup{} refer to the cases where only supervised loss is used. ICT[K] refer to a setup where the ICT loss with $K$ \mixup{} points are used as unsupervised loss. The supervised loss with the ICT does not use \mixup{}.}
\label{fig:cifar10_labels250_cw100_linearity}
\end{figure*}

\paragraph{Results} 
Figure~\ref{fig:cifar10_labels250_cw100_linearity} show the evolution of non-linearity throughout network training for different baselines on CIFAR-10. These figures use a labeled dataset of size $250$ (see Appendix for other datasets and settings). The total training iterations are remapped between $0$ and $1000$ and referred to as \texttt{Train time} in the plots.  

First, we compare the (non-)linearity of ERM and \mixup{} (both are supervised setups) in different layers. We see that in the first and last layer the representations from \mixup{} are more linear compared to ERM representations. Intuitively, this could be explained by the \mixup{} operation linearly combining both the inputs and outputs. Hence, training using \mixup{} keeps the first and last functions in the composition $f = f_0 \circ f_1 \cdots \circ f_5$ more linear. Surprisingly, however, we do not see that \mixup{} representations remain more linear in \emph{all} the layers compared to those of ERM as suggested by the theory. For example, in Figure~\ref{fig:cifar10_labels250_cw100_linearity}, we can see that \texttt{Layers 2, 3, 4} based on \mixup{} training are more non-linear compared to those trained with ERM. This suggests that a) the findings of theory on simplified setups do not directly translate to practical deep networks, and b) more importantly, without proper regularization in all layers, the network maintains an overall non-linearity from input to output. Further, this suggests that the linearity enforced by a regularization in some layers is counteracted by increased non-linearity in the other layers. 

Second, we see that ICT tend to make \emph{all} the layers more linear compared to ERM and \mixup{}, except for the first layer where \mixup{} is slightly more linear. This suggests that the ICT loss, as defined in \eqref{eq:ictx}, applied on the network logits, and on a larger unsupervised dataset, is a stronger regularization compared to \mixup{} defined on one-hot encoded vectors. Further, we see from Figure~\ref{fig:cifar10_labels250_cw100_linearity} that using more \mixup{} points (ICT4 vs ICT2), leads to stronger linearity across all layers. In the next section, we will see that this stronger linearity also leads to faster convergence.

\subsection{Effect of stronger \mixup{} on convergence} \label{sec:stronger_mixup}
In the previous section, we saw that stronger enforcement of linearity, by using more \mixup{} points, led to the network layers becoming more linear. We will now empirically show the effect of this linearity on network convergence. Specifically, we measure test accuracy given a specified number of labeled training examples. For this, we will investigate the performance of different baselines in a semi-supervised setup for varying amounts of labeled data. 

\paragraph{Results}

\begin{figure}[t]
  \begin{center}
    \includegraphics[width=0.48\textwidth]{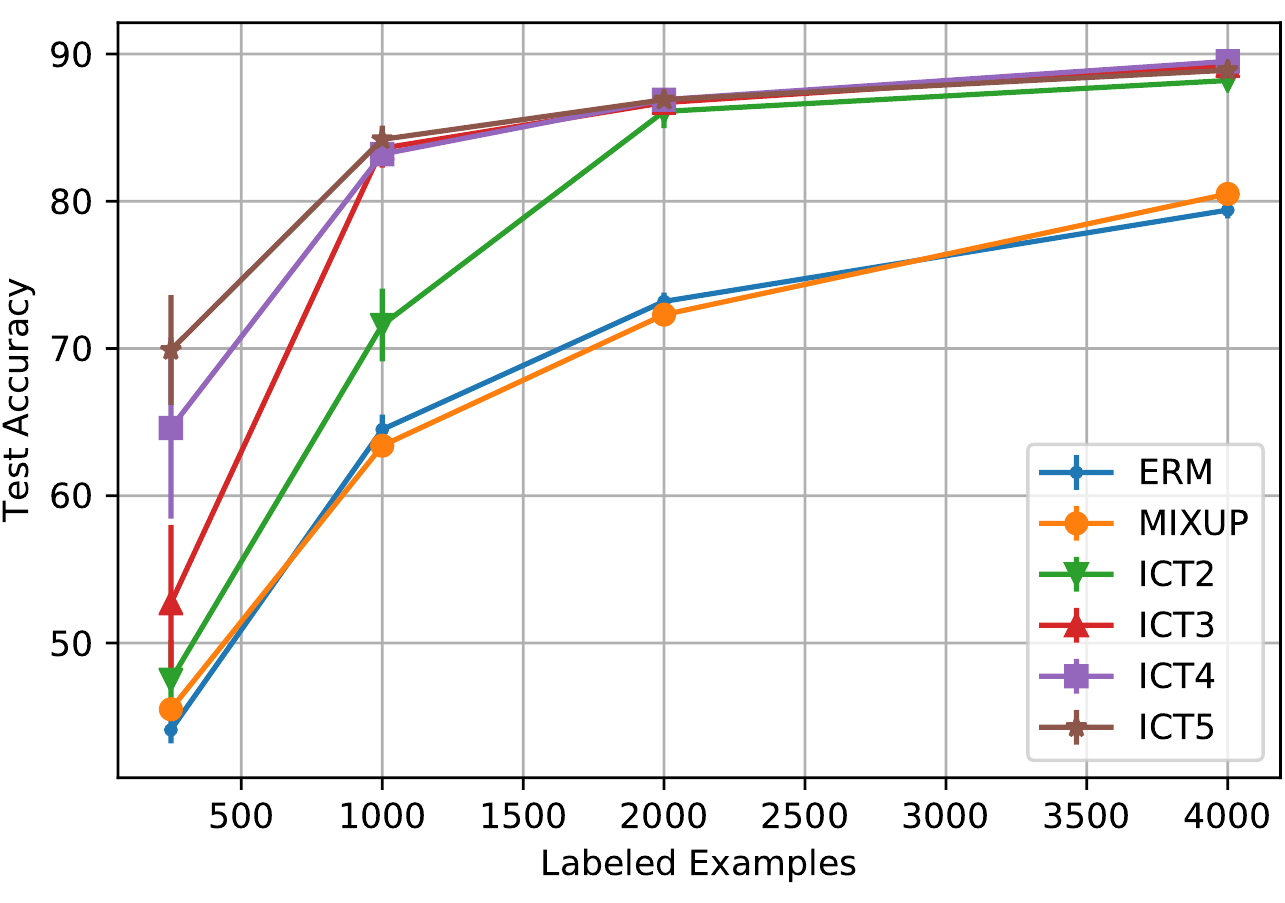}
  \end{center}
  \negspace{-1mm}
  \caption{\textbf{CIFAR-10}: Test accuracy with varying numbers of labeled training examples.
  ERM and \mixup{} refer to the cases where only supervised loss is used. ICT[K] refer to a setup where the ICT loss with K \mixup{} points are used as unsupervised loss. The supervised loss with the ICT does not use \mixup{} loss.}
  \label{fig:cifar10_ictX}
\end{figure}

Figure~\ref{fig:cifar10_ictX} shows the accuracy on a fixed test set against various labeled dataset sizes for CIFAR-10 (the results for CIFAR-100 are given in the appendix). From the Figure, we can see that the ICT-based model generally outperforms the ERM and \mixup{} baselines across all dataset sizes. This is likely due to the additional unsupervised training performed on the additional unsupervised dataset. Among the ICT variants we further observe that the test accuracy tends to be higher as we increase the number of points used in the \mixup{} operation. This is especially true in the regime of very small datasets and the performance of all ICT variants converges to similar performance as more datapoints become available. For example, to reach $70$\% test accuracy on CIFAR-10, \mixup{} requires roughly $1800$ labeled examples, whereas ICT4 only requires $500$ labeled examples (even less for ICT5). The stronger regularization achieved by using 4 \mixup{} points leads to the network becoming more linear, which results in more efficient learning from fewer labeled samples. 

\section{Conclusion} \label{sec:conc}

In this work, we explored when does (or doesn't) \mixup{} enforce local linearity in the learned representations.
We studied this question in a semi-supervised learning setup for image classification tasks.
Our experiments demonstrate that standard supervised \mixup{} training doesn't make the representations in all the layers locally linear.
In fact, some of the intermediate layers become more non-linear during supervised \mixup{} training compared to standard empirical risk minimization (ERM) training. 
However, when \mixup{} operation is used as an unsupervised loss, on a larger unlabeled dataset, network representations in all layers become more linear.
As a consequence of these smoother representations, the network converges to a given test set performance faster (in terms of labeled samples) than a network that is locally less linear.
These findings, which we verified empirically, sheds new light on the training dynamics of a network trained under \mixup{} regularization.
\clearpage
{\small \bibliography{main}}
\bibliographystyle{plainnat}

\section*{Checklist}

\begin{enumerate}

\item For all authors...
\begin{enumerate}
  \item Do the main claims made in the abstract and introduction accurately reflect the paper's contributions and scope?
    \answerYes{}
  \item Did you describe the limitations of your work?
    \answerYes{Please see \secref{sec:intro}.}
  \item Did you discuss any potential negative societal impacts of your work?
    \answerNA{}
  \item Have you read the ethics review guidelines and ensured that your paper conforms to them?
    \answerYes{}
\end{enumerate}

\item If you are including theoretical results...
\begin{enumerate}
  \item Did you state the full set of assumptions of all theoretical results?
    \answerNA{}
        \item Did you include complete proofs of all theoretical results?
    \answerNA{}
\end{enumerate}

\item If you ran experiments...
\begin{enumerate}
  \item Did you include the code, data, and instructions needed to reproduce the main experimental results (either in the supplemental material or as a URL)?
    \answerYes{Please see \secref{sec:experiments}.}
  \item Did you specify all the training details (e.g., data splits, hyperparameters, how they were chosen)?
    \answerYes{Please see \secref{sec:experiments}.}
  \item Did you report error bars (e.g., with respect to the random seed after running experiments multiple times)? \answerYes{}
  \item Did you include the total amount of compute and the type of resources used (e.g., type of GPUs, internal cluster, or cloud provider)? \answerNo{}
\end{enumerate}

\item If you are using existing assets (e.g., code, data, models) or curating/releasing new assets...
\begin{enumerate}
  \item If your work uses existing assets, did you cite the creators?
    \answerYes{}
  \item Did you mention the license of the assets?
    \answerYes{}
  \item Did you include any new assets either in the supplemental material or as a URL?
    \answerNo{}
  \item Did you discuss whether and how consent was obtained from people whose data you're using/curating?
    \answerNA{}
  \item Did you discuss whether the data you are using/curating contains personally identifiable information or offensive content?
    \answerNA{}
\end{enumerate}

\item If you used crowdsourcing or conducted research with human subjects...
\begin{enumerate}
  \item Did you include the full text of instructions given to participants and screenshots, if applicable?
    \answerNA{}
  \item Did you describe any potential participant risks, with links to Institutional Review Board (IRB) approvals, if applicable?
    \answerNA{}
  \item Did you include the estimated hourly wage paid to participants and the total amount spent on participant compensation?
    \answerNA{}
\end{enumerate}

\end{enumerate}

\clearpage
\newpage
\onecolumn
\section*{Appendix}

\appendix

In this appendix we provide more detailed results for various sizes of labeled dataset. The results here further strengthen the claims made in the main paper.  

\section{More Results} \label{sec:sup_more_results}

\begin{figure*}[t]
\def \LABELS {2000}
\def \CTWEIGHT {100}
\centering
\begin{subfigure}{.33\textwidth}
      \centering
      \includegraphics[width=.99\linewidth]{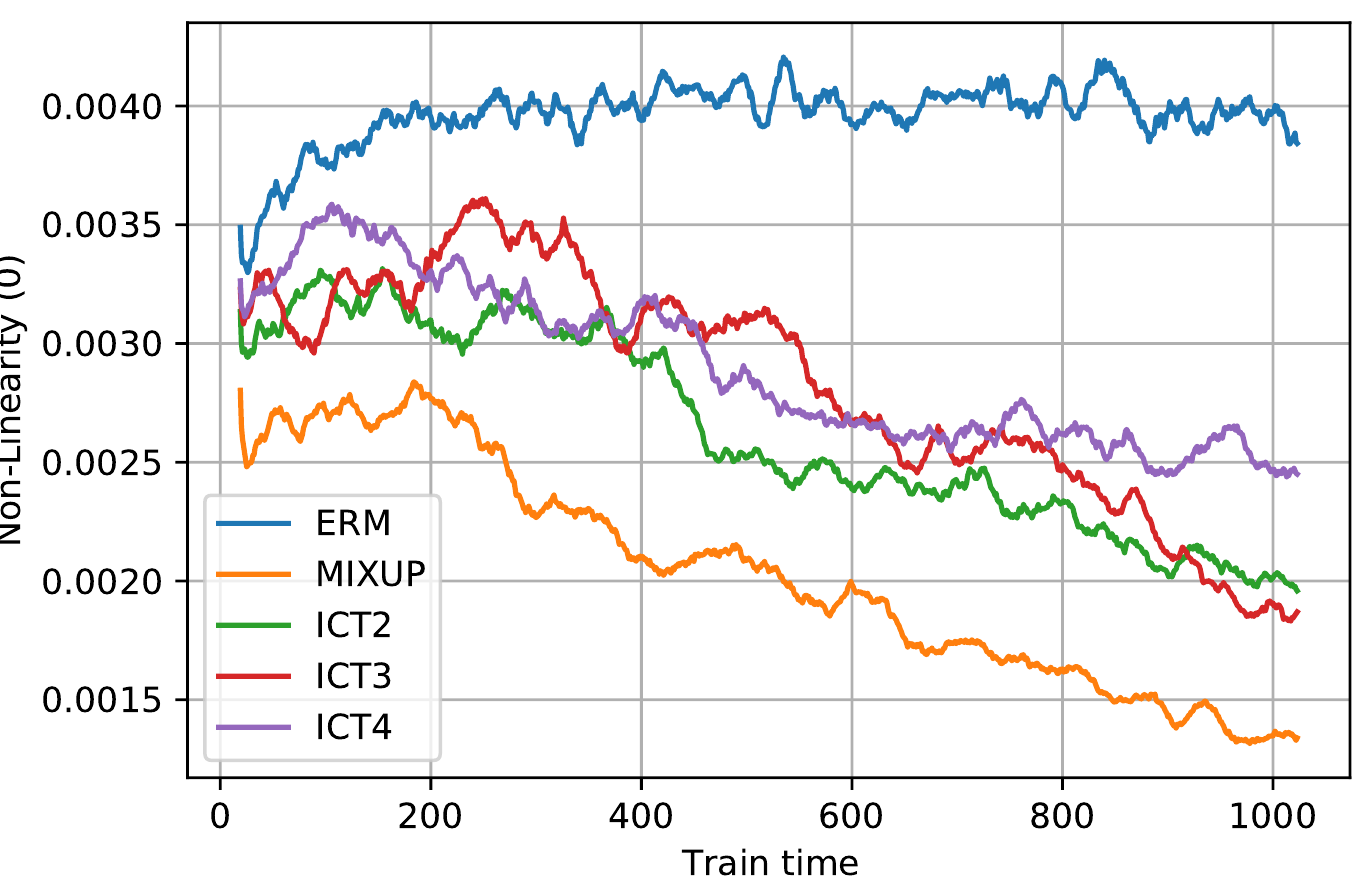}
      \caption{\small Layer 0}
\end{subfigure}\hfill
\begin{subfigure}{.33\textwidth}
      \centering
      \includegraphics[width=.99\linewidth]{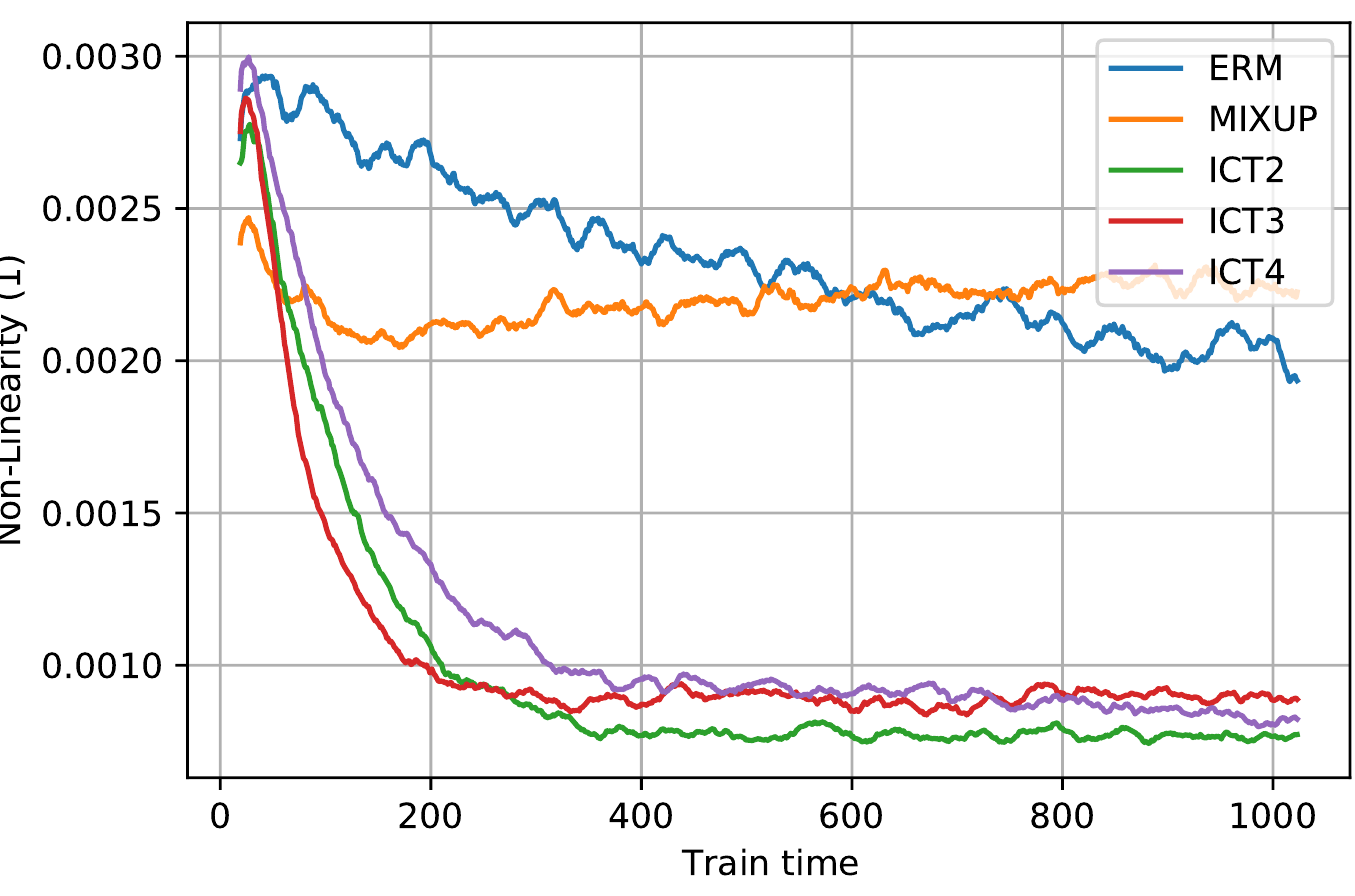}
      \caption{\small Layer 1}
\end{subfigure}\hfill
\begin{subfigure}{.33\textwidth}
      \centering
      \includegraphics[width=.99\linewidth]{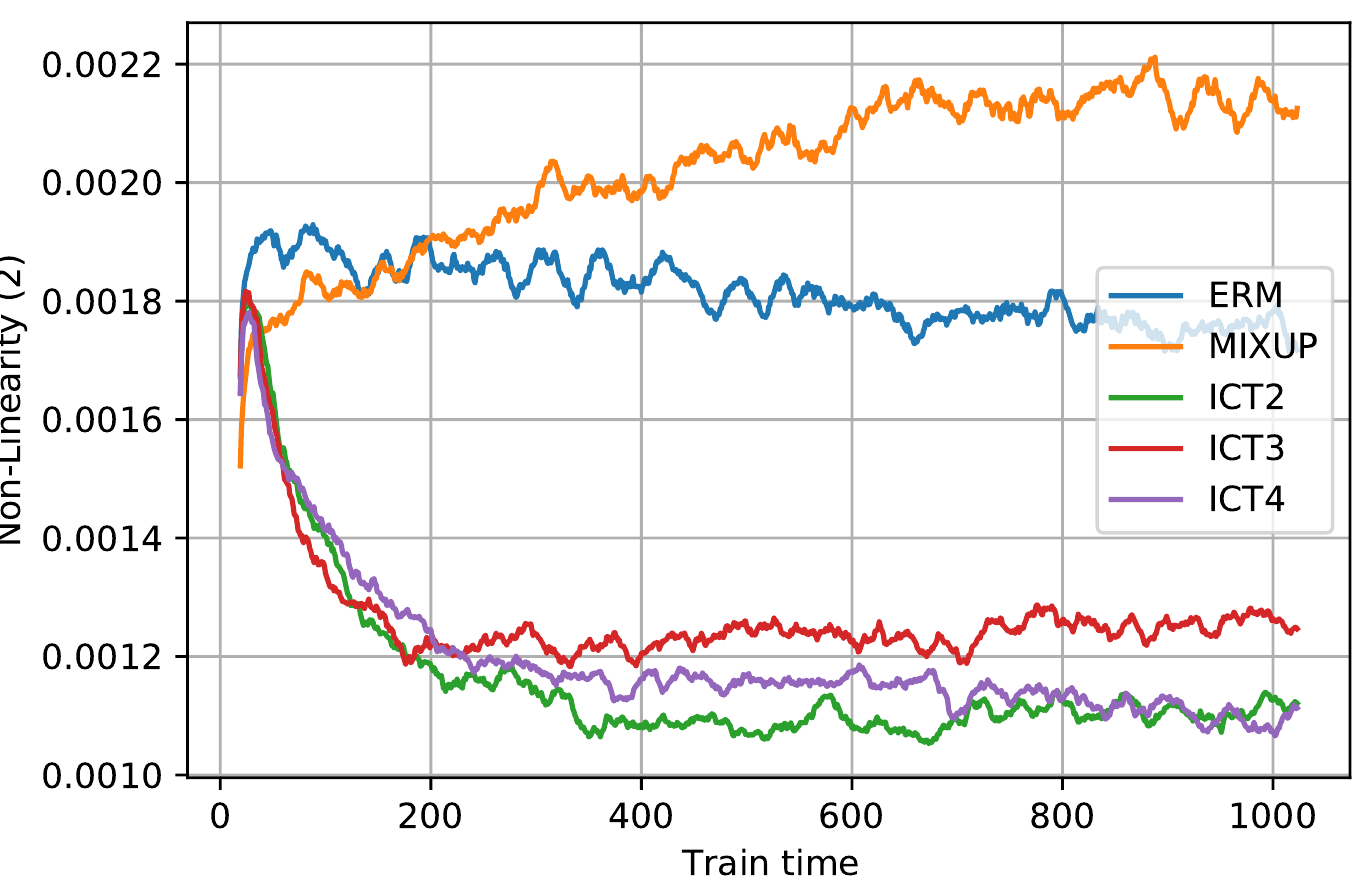}
      \caption{\small Layer 2}
\end{subfigure}

\begin{subfigure}{.33\textwidth}
      \centering
      \includegraphics[width=.99\linewidth]{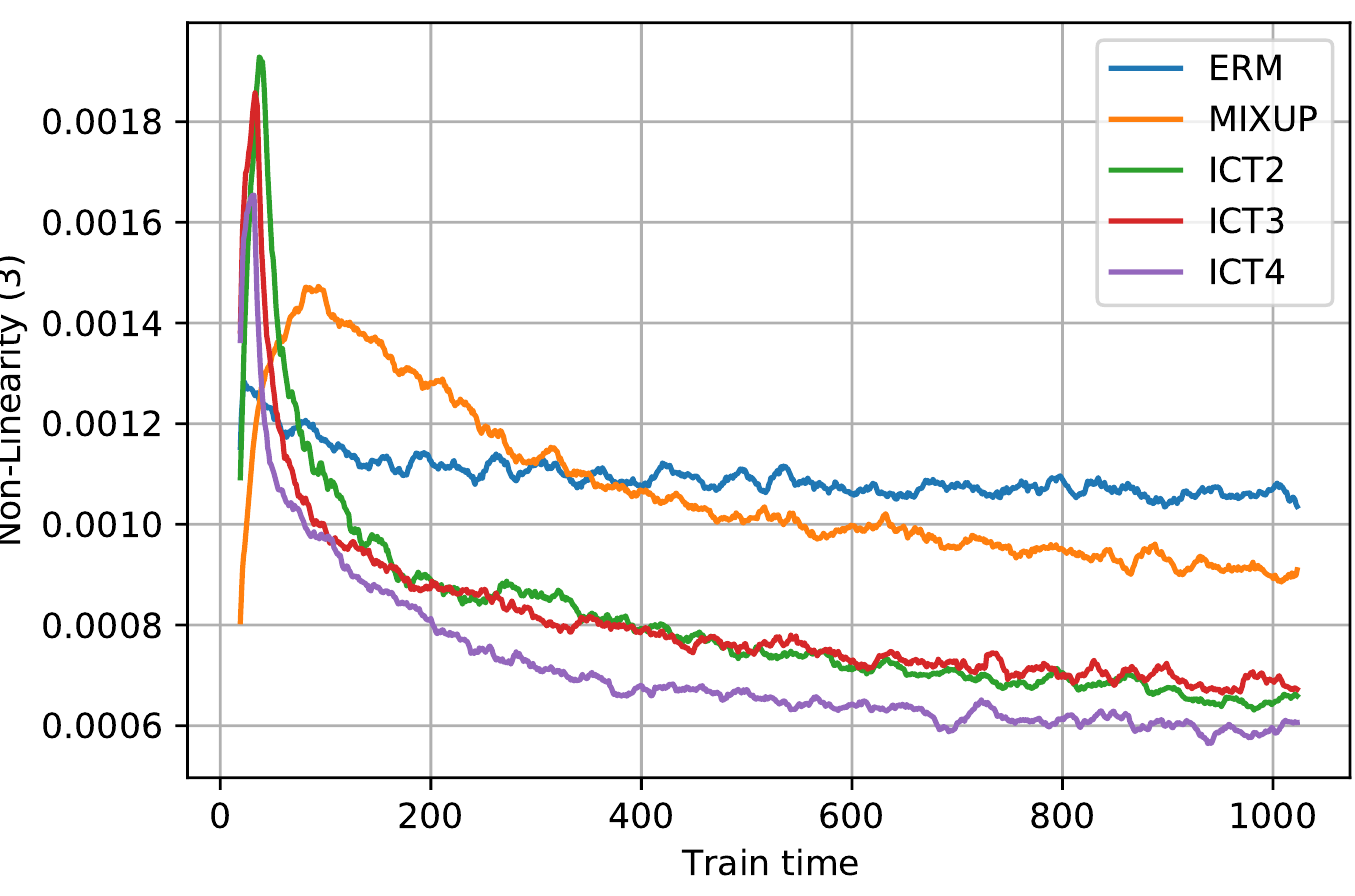}
      \caption{\small Layer 3}
\end{subfigure}\hfill
\begin{subfigure}{.33\textwidth}
      \centering
      \includegraphics[width=.99\linewidth]{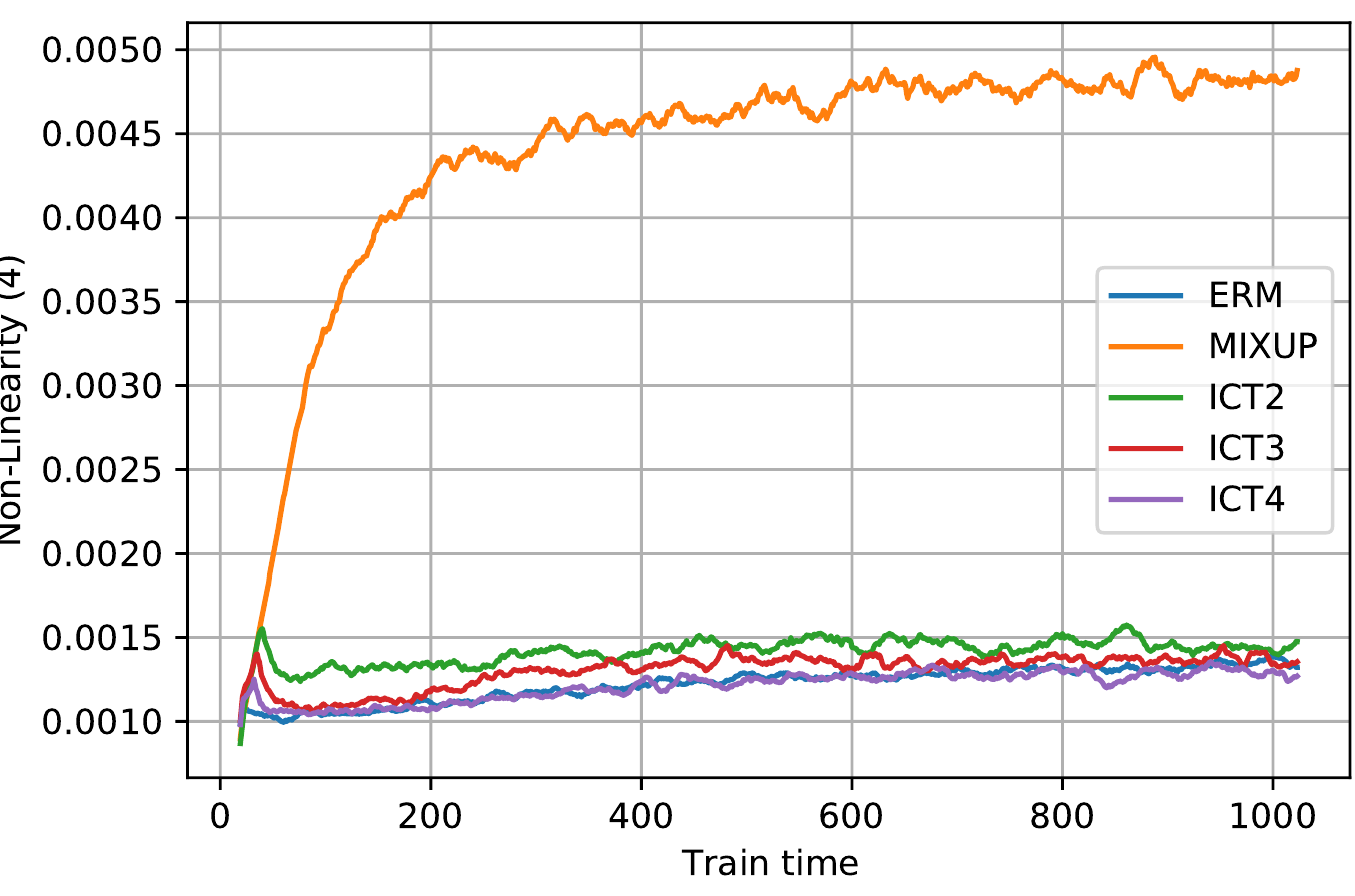}
      \caption{\small Layer 4}
\end{subfigure}\hfill
\begin{subfigure}{.33\textwidth}
      \centering
      \includegraphics[width=.99\linewidth]{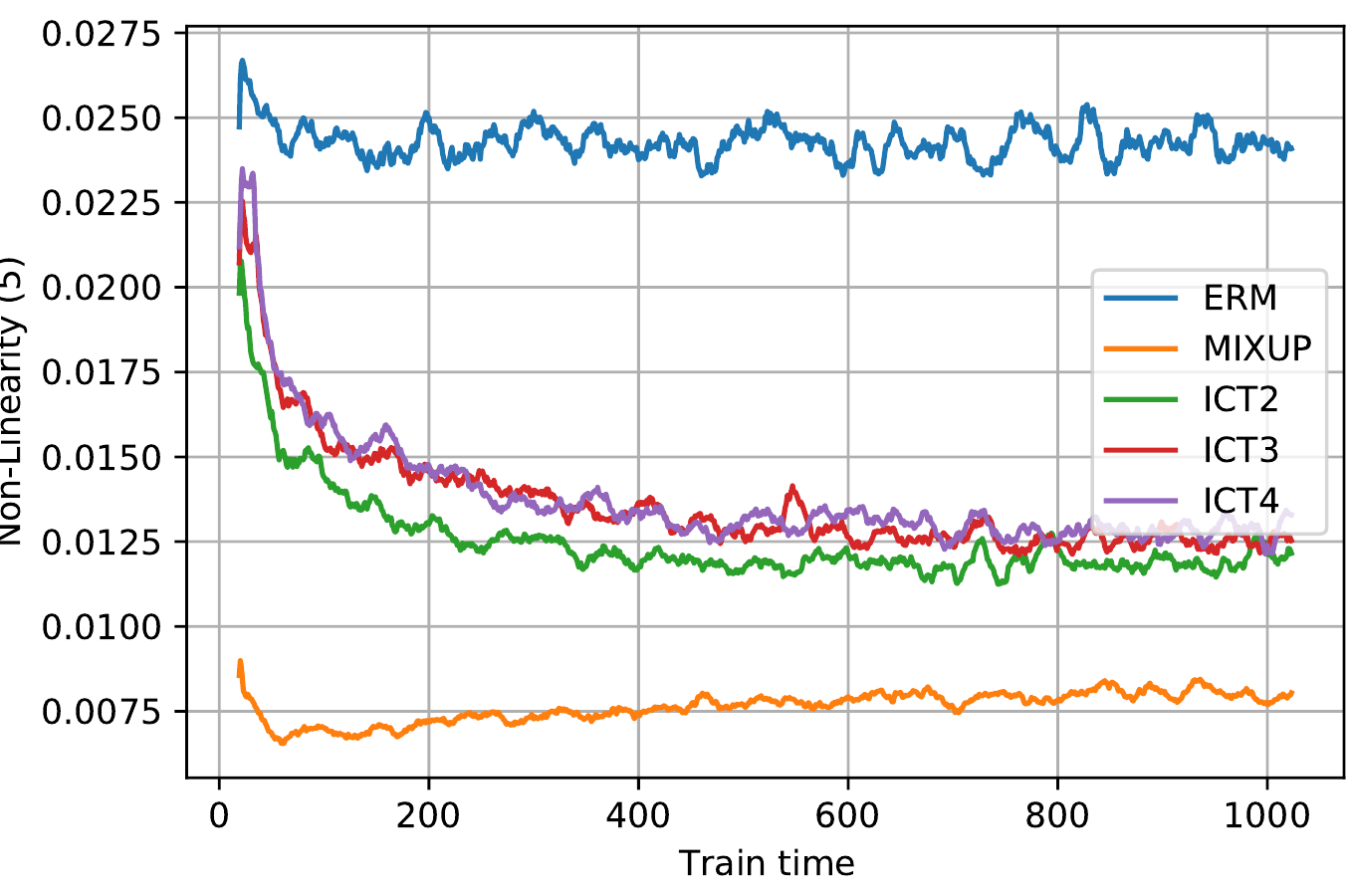}
      \caption{\small Layer 5}
\end{subfigure}

\negspace{-1mm}
\caption{\textbf{CIFAR-10}: Linearity propagation in different layers with \LABELS\ labeled examples and consistency weight of \CTWEIGHT. ERM and mixup refer to the cases where only supervised loss is used. $\ictx_K$ refer to a setup where the ICT loss with $K$ mixup points are used as unsupervised loss. The supervised loss with the ICT does not use mixup.}
\label{fig:cifar10_labels2000_cw100_linearity}
\negspace{-3mm}
\end{figure*}

\begin{figure*}[t]
\def \LABELS {4000}
\def \CTWEIGHT {100}
\centering
\begin{subfigure}{.33\textwidth}
      \centering
      \includegraphics[width=.99\linewidth]{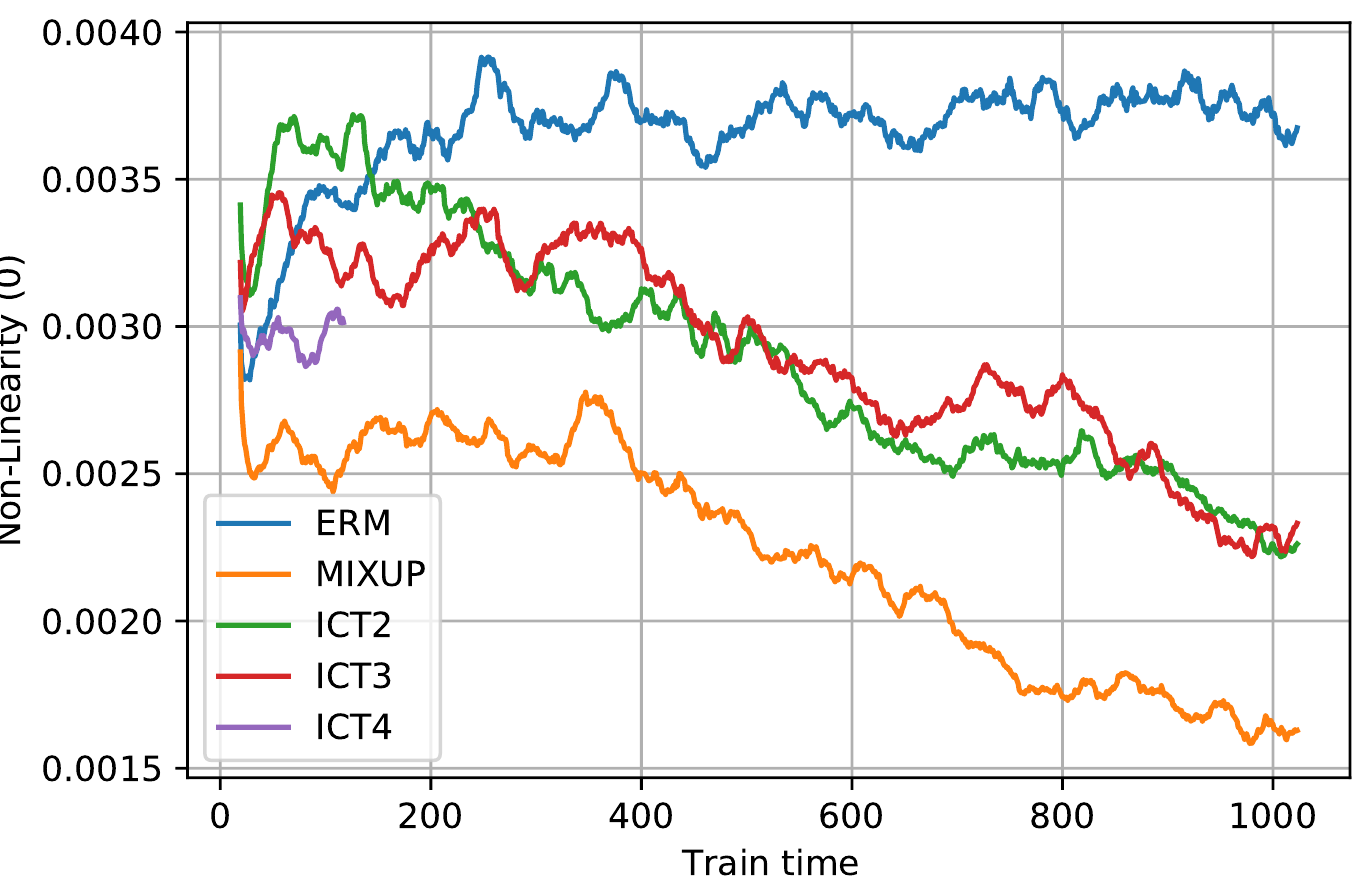}
      \caption{\small Layer 0}
\end{subfigure}\hfill
\begin{subfigure}{.33\textwidth}
      \centering
      \includegraphics[width=.99\linewidth]{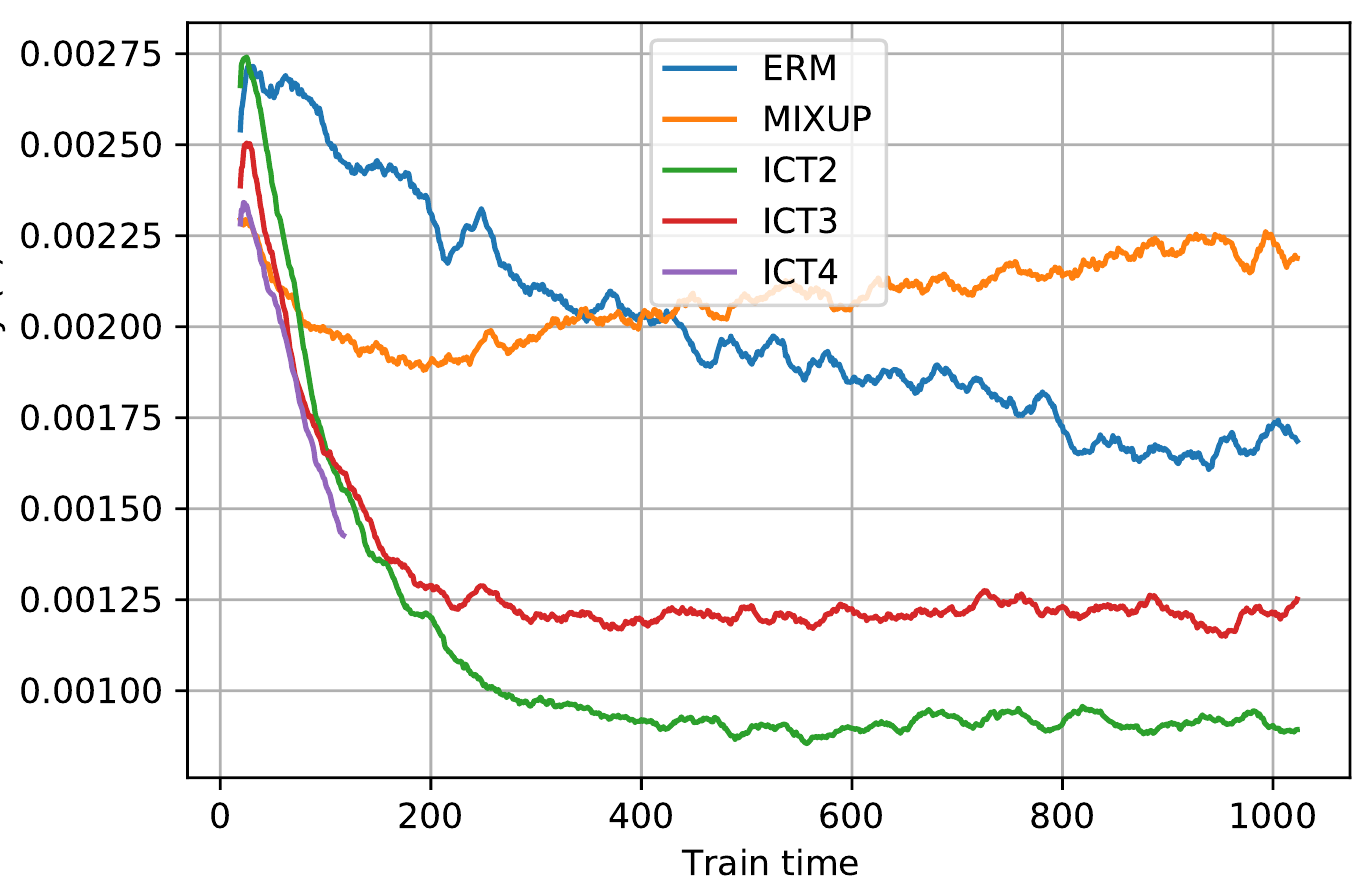}
      \caption{\small Layer 1}
\end{subfigure}\hfill
\begin{subfigure}{.33\textwidth}
      \centering
      \includegraphics[width=.99\linewidth]{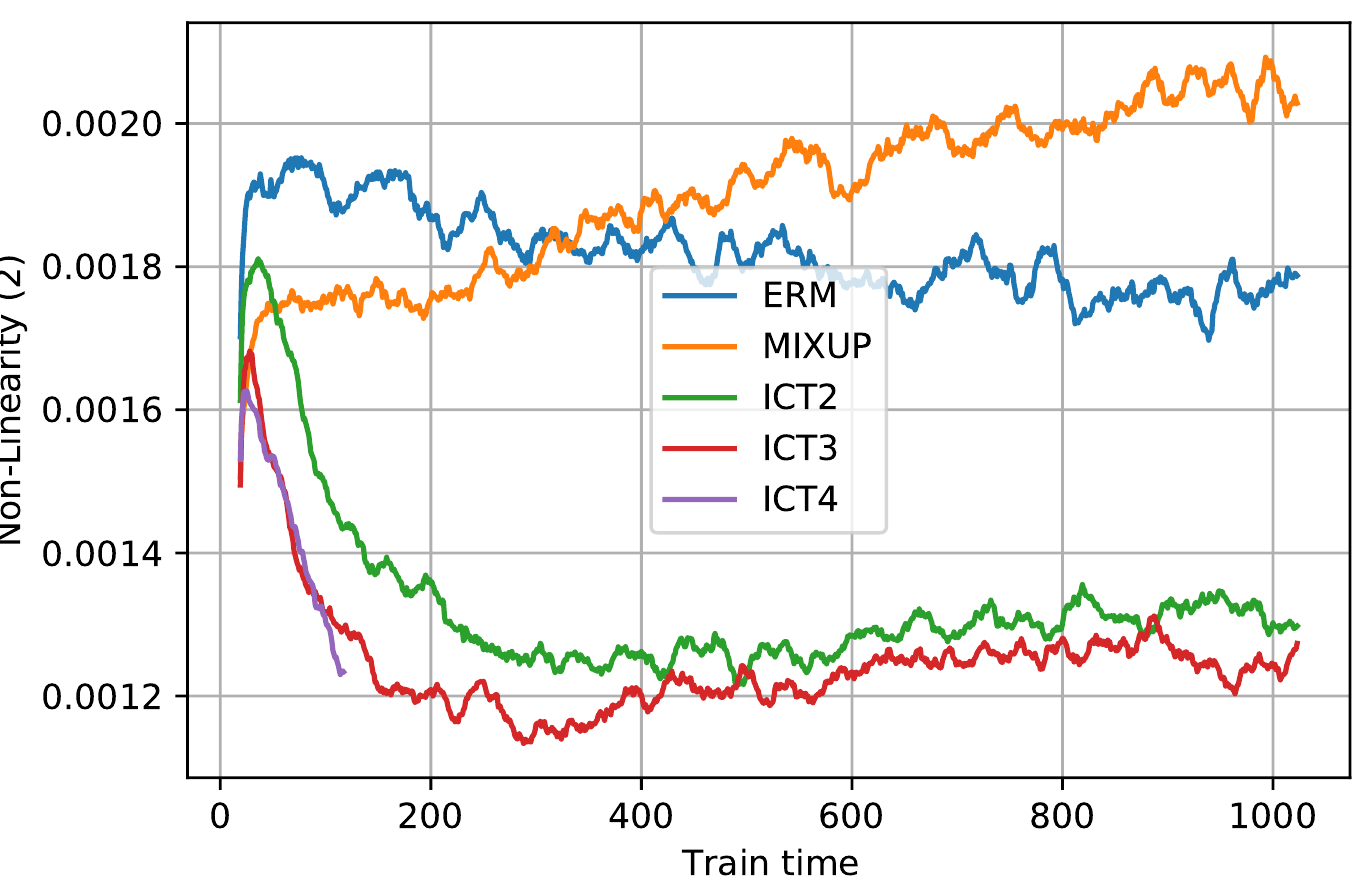}
      \caption{\small Layer 2}
\end{subfigure}

\begin{subfigure}{.33\textwidth}
      \centering
      \includegraphics[width=.99\linewidth]{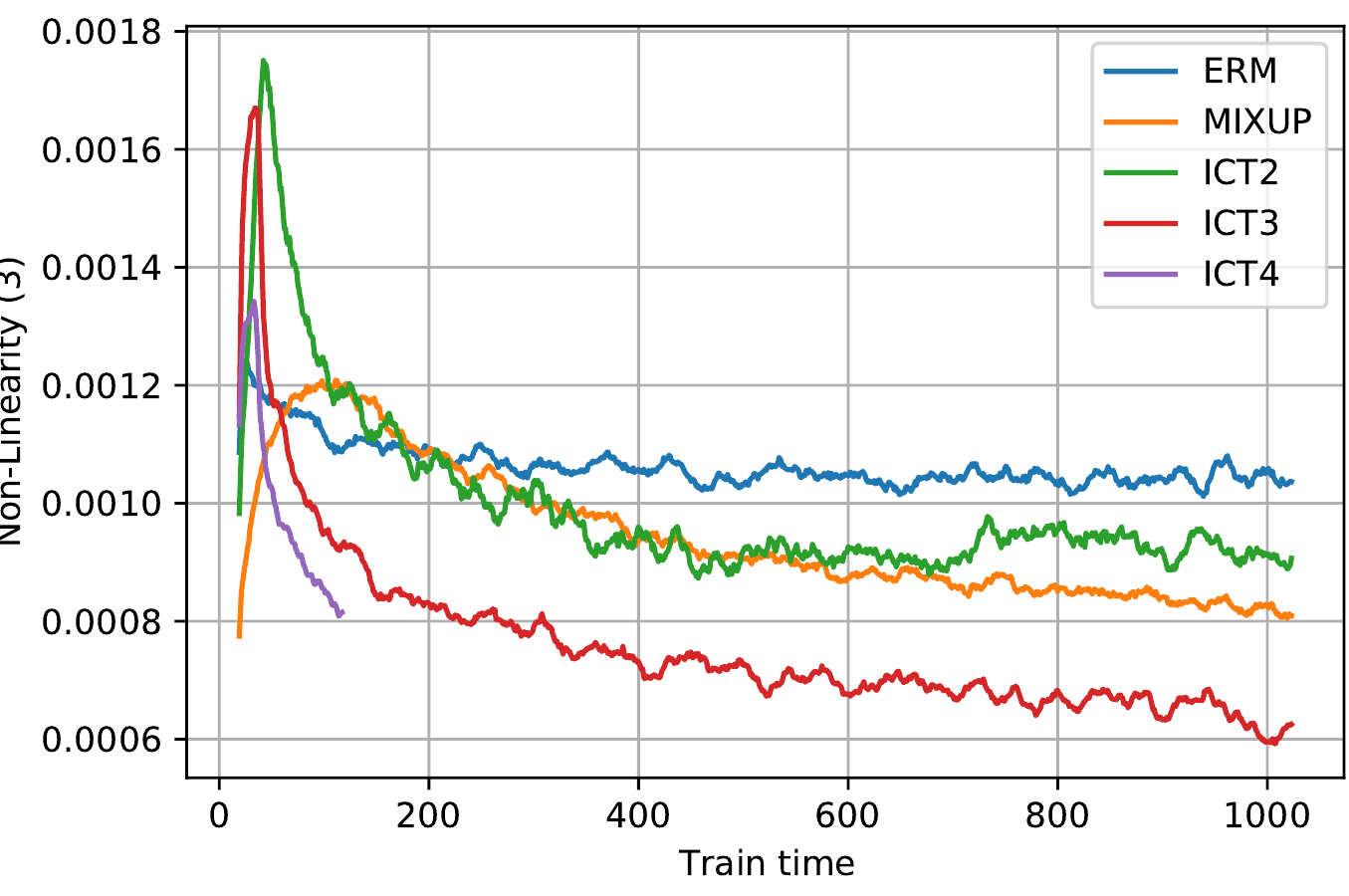}
      \caption{\small Layer 3}
\end{subfigure}\hfill
\begin{subfigure}{.33\textwidth}
      \centering
      \includegraphics[width=.99\linewidth]{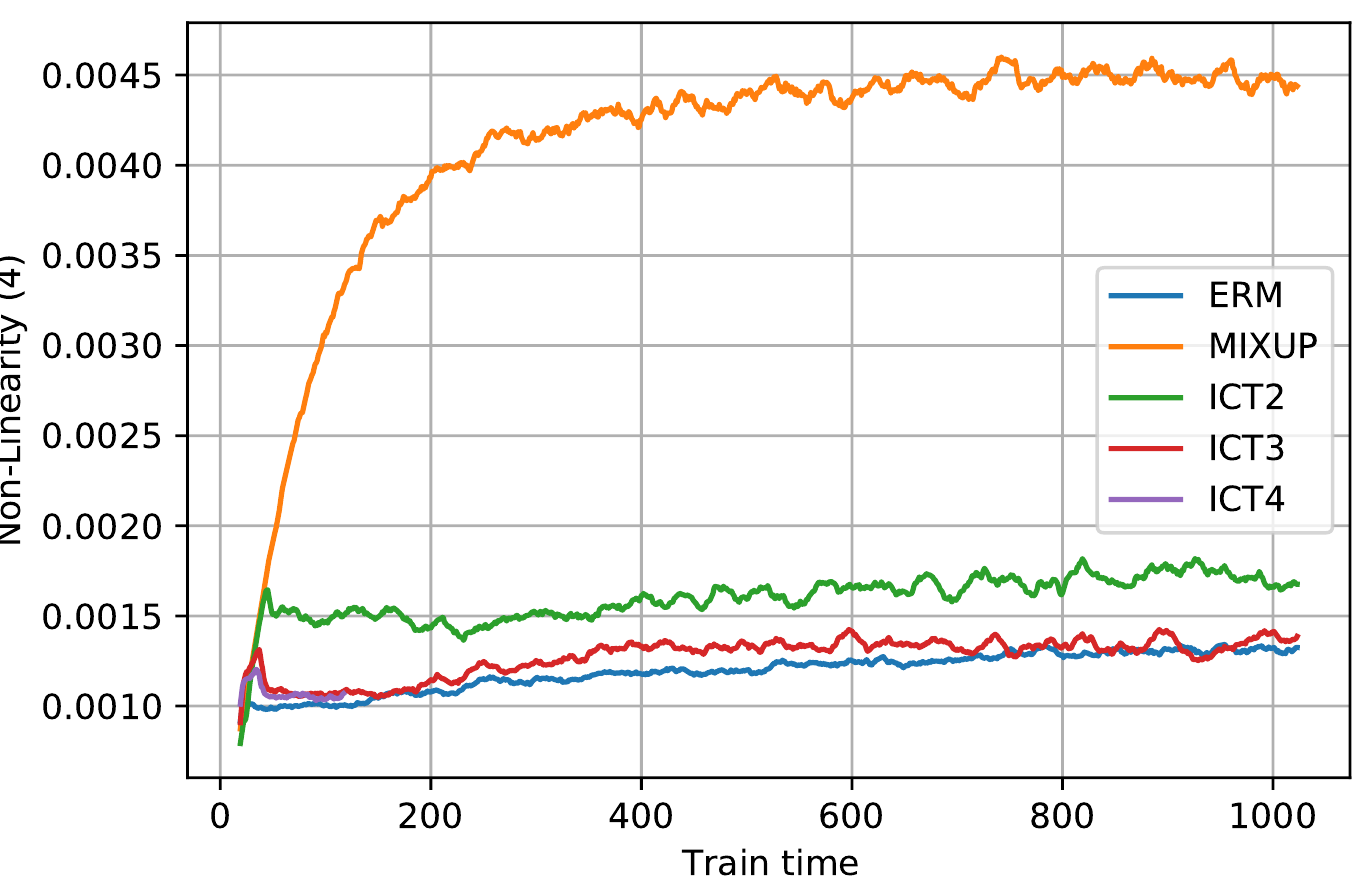}
      \caption{\small Layer 4}
\end{subfigure}\hfill
\begin{subfigure}{.33\textwidth}
      \centering
      \includegraphics[width=.99\linewidth]{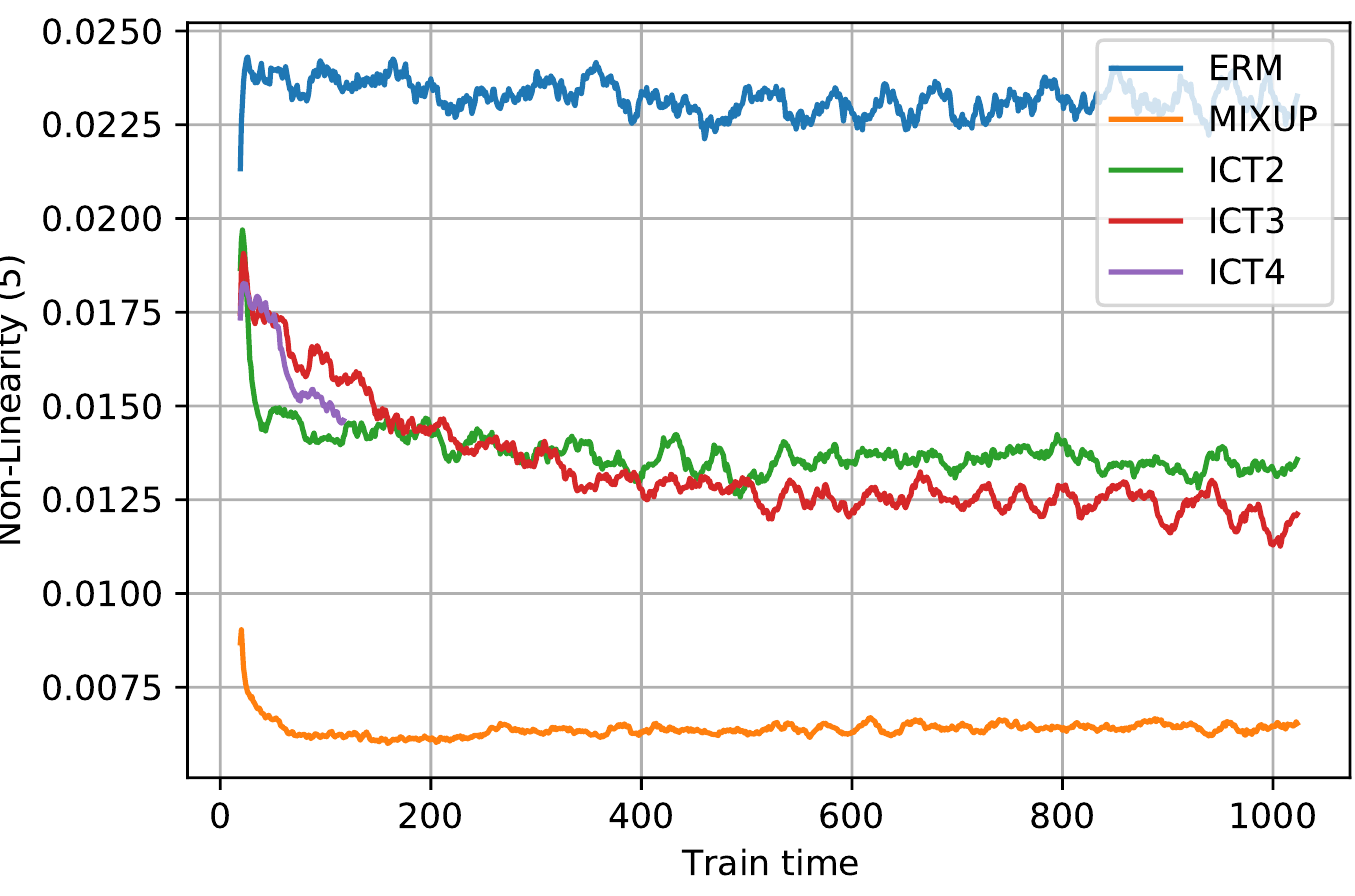}
      \caption{\small Layer 5}
\end{subfigure}

\negspace{-1mm}
\caption{\textbf{CIFAR-10}: Linearity propagation in different layers with \LABELS\ labeled examples and consistency weight of \CTWEIGHT. ERM and mixup refer to the cases where only supervised loss is used. $\ictx_K$ refer to a setup where the ICT loss with $K$ mixup points are used as unsupervised loss. The supervised loss with the ICT does not use mixup.}
\label{fig:cifar10_labels4000_cw100_linearity}
\negspace{-3mm}
\end{figure*}

\begin{figure*}[t]
\def \LABELS {250}
\def \CTWEIGHT {100}	
\centering
\begin{subfigure}{.33\textwidth}
      \centering
      \includegraphics[width=.99\linewidth]{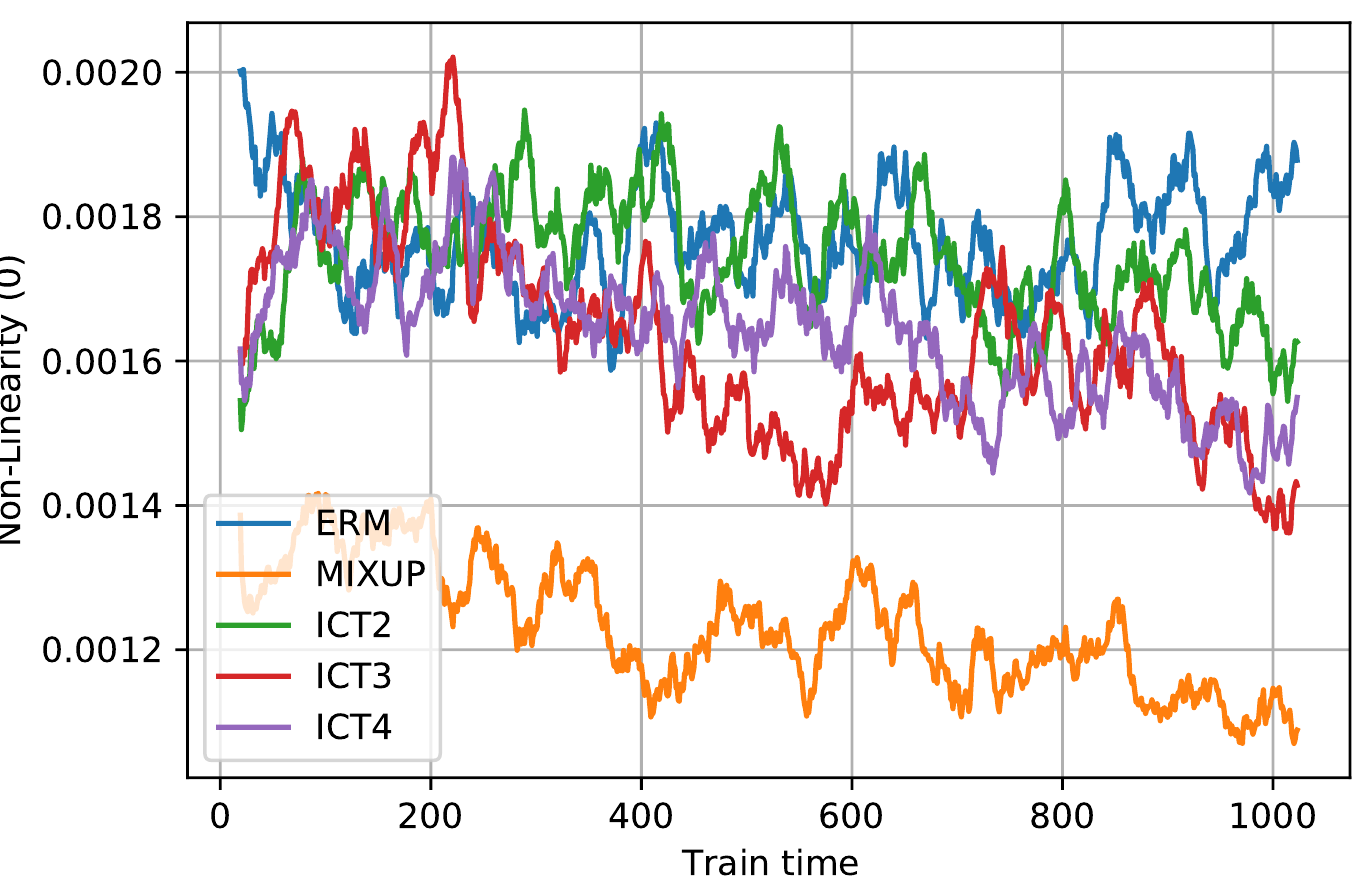}
      \caption{\small Layer 0}
\end{subfigure}\hfill
\begin{subfigure}{.33\textwidth}
      \centering
      \includegraphics[width=.99\linewidth]{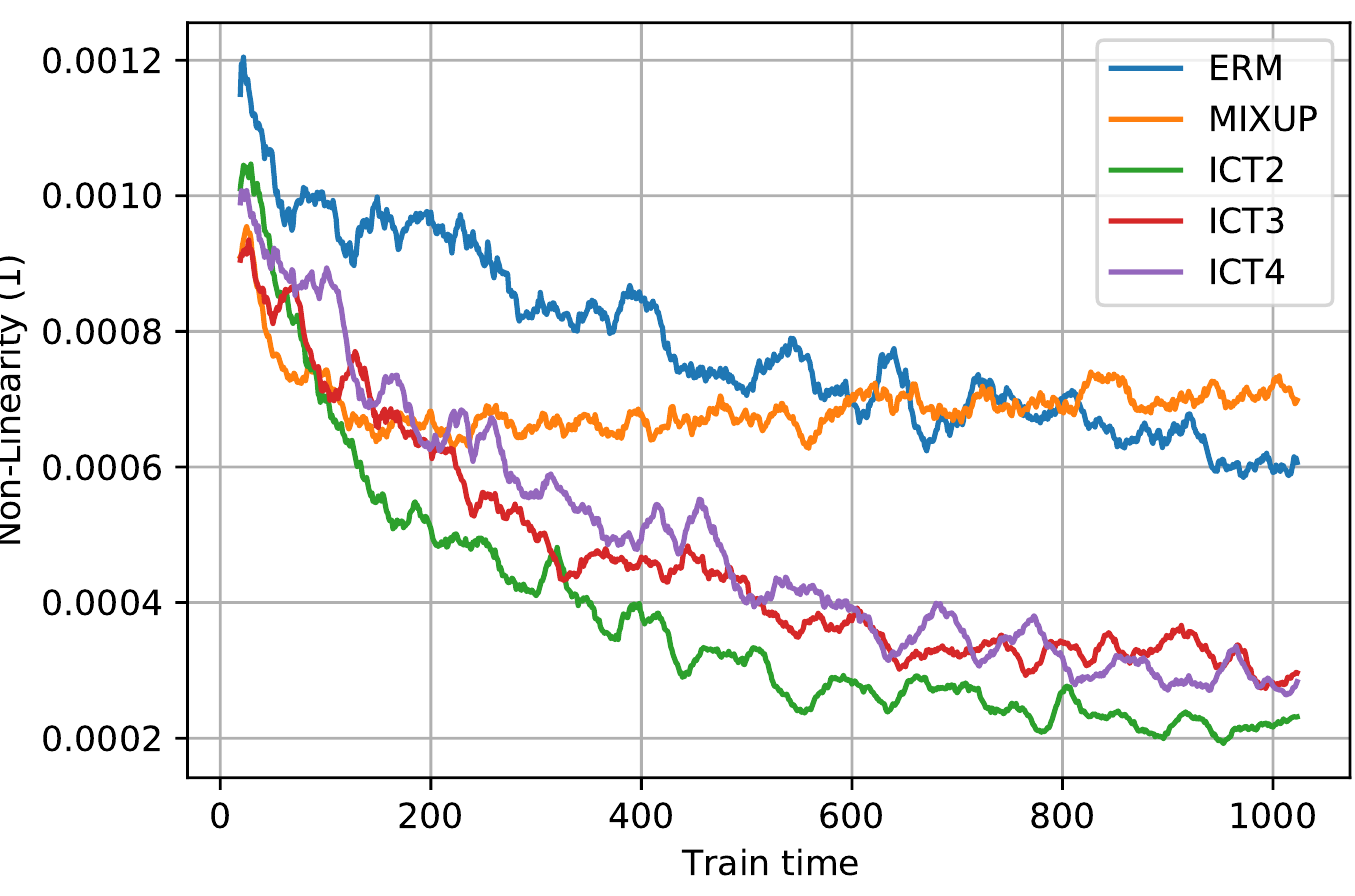}
      \caption{\small Layer 1}
\end{subfigure}\hfill
\begin{subfigure}{.33\textwidth}
      \centering
      \includegraphics[width=.99\linewidth]{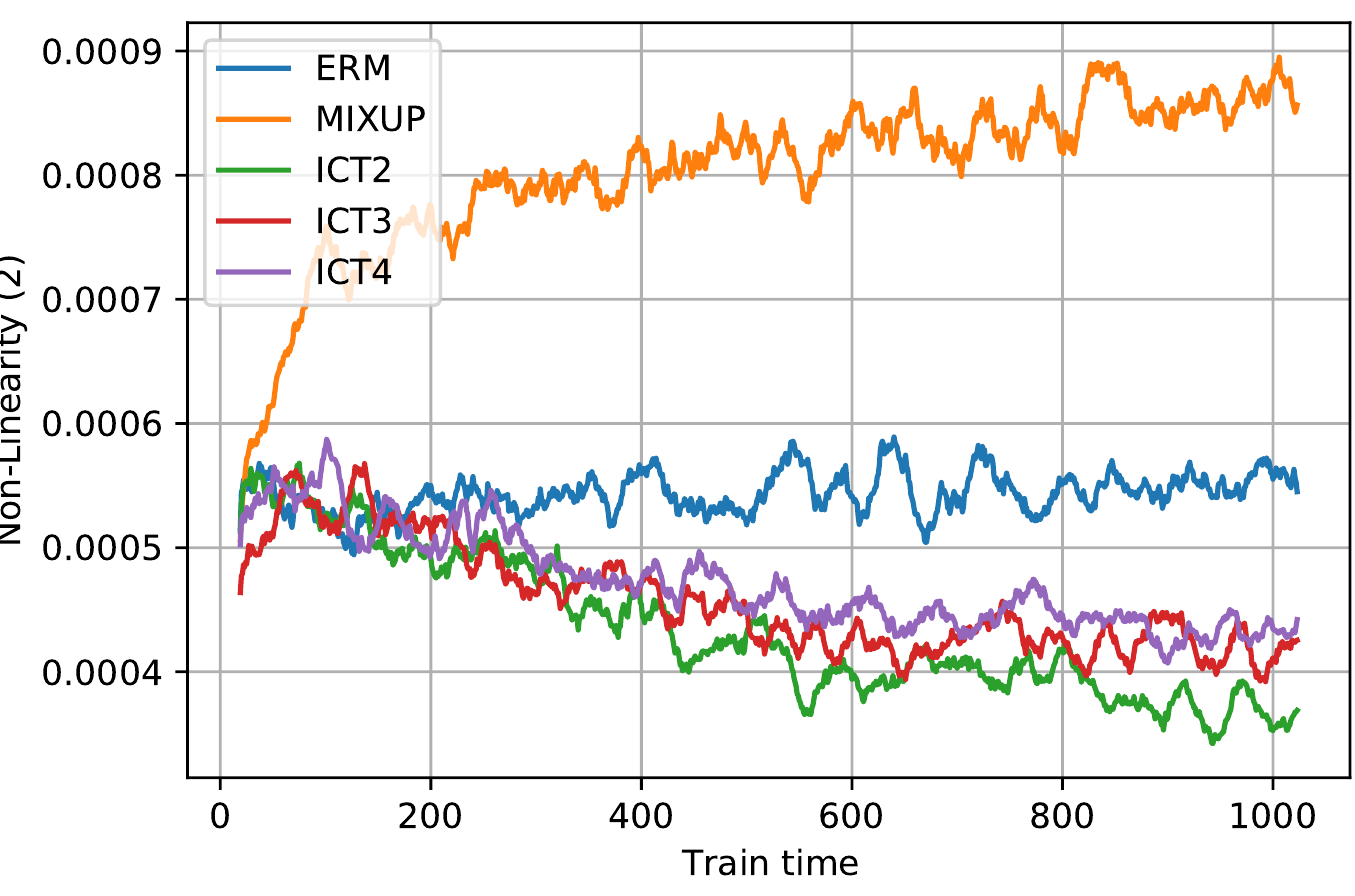}
      \caption{\small Layer 2}
\end{subfigure}

\begin{subfigure}{.33\textwidth}
      \centering
      \includegraphics[width=.99\linewidth]{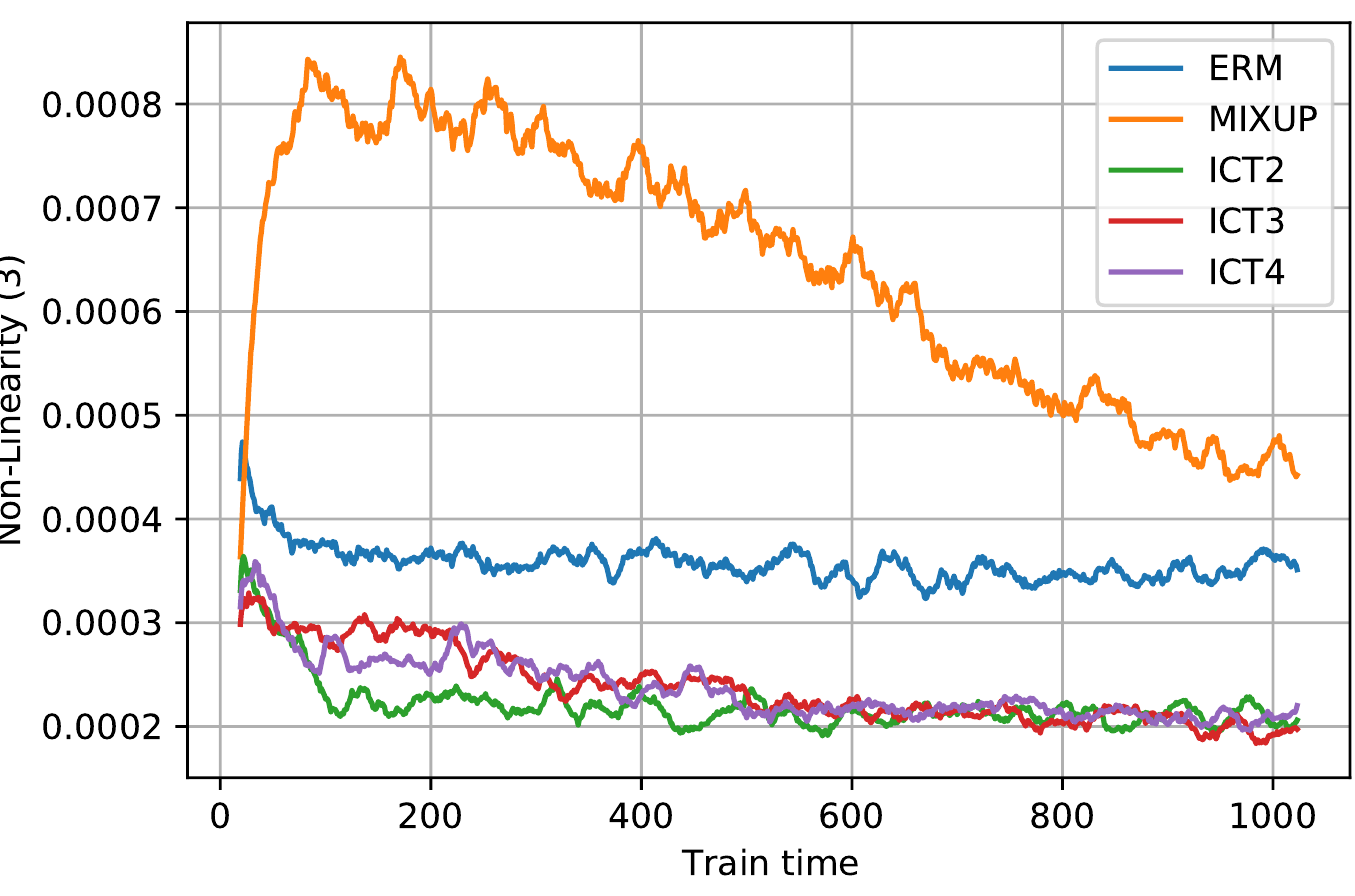}
      \caption{\small Layer 3}
\end{subfigure}\hfill
\begin{subfigure}{.33\textwidth}
      \centering
      \includegraphics[width=.99\linewidth]{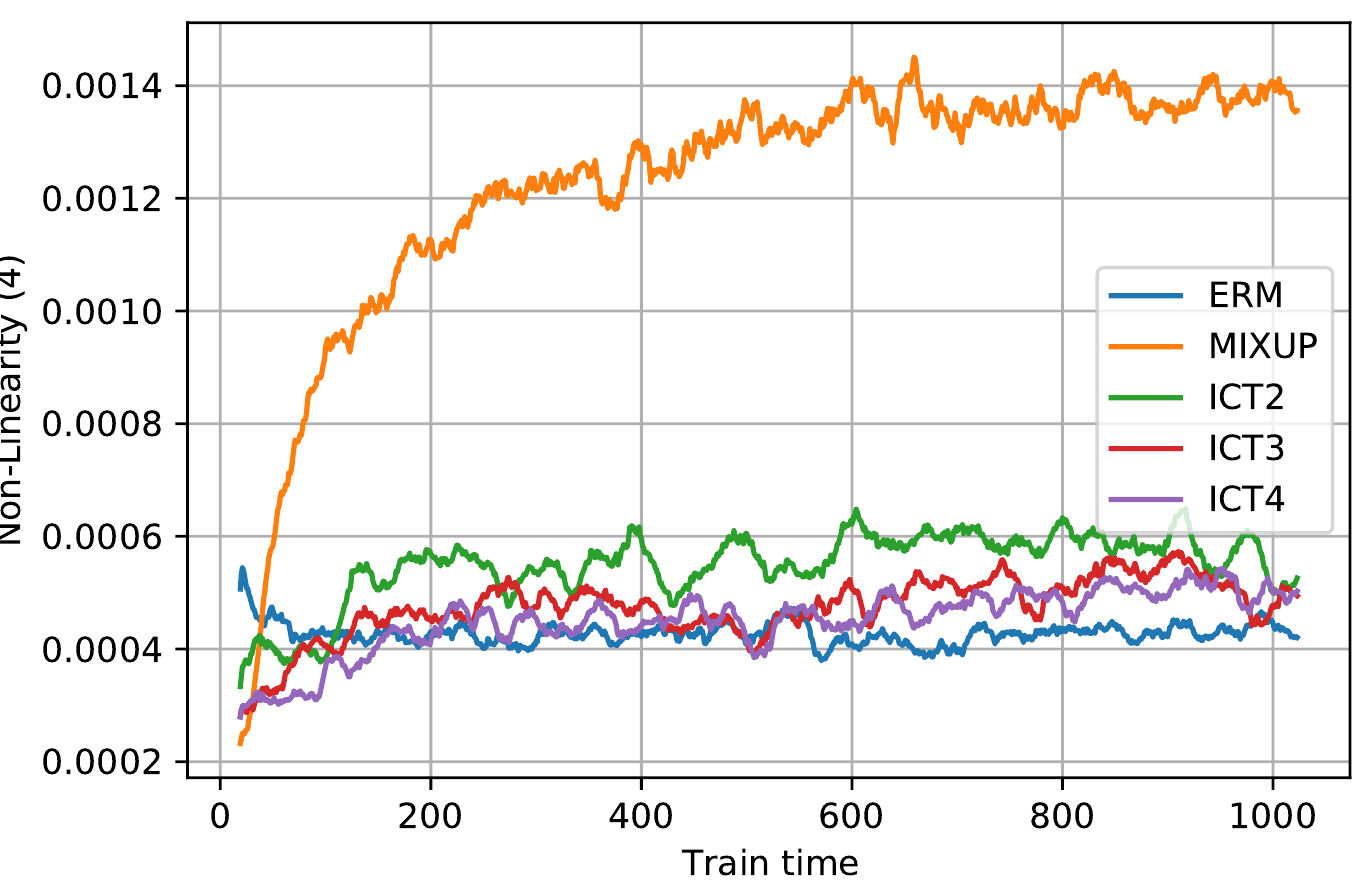}
      \caption{\small Layer 4}
\end{subfigure}\hfill
\begin{subfigure}{.33\textwidth}
      \centering
      \includegraphics[width=.99\linewidth]{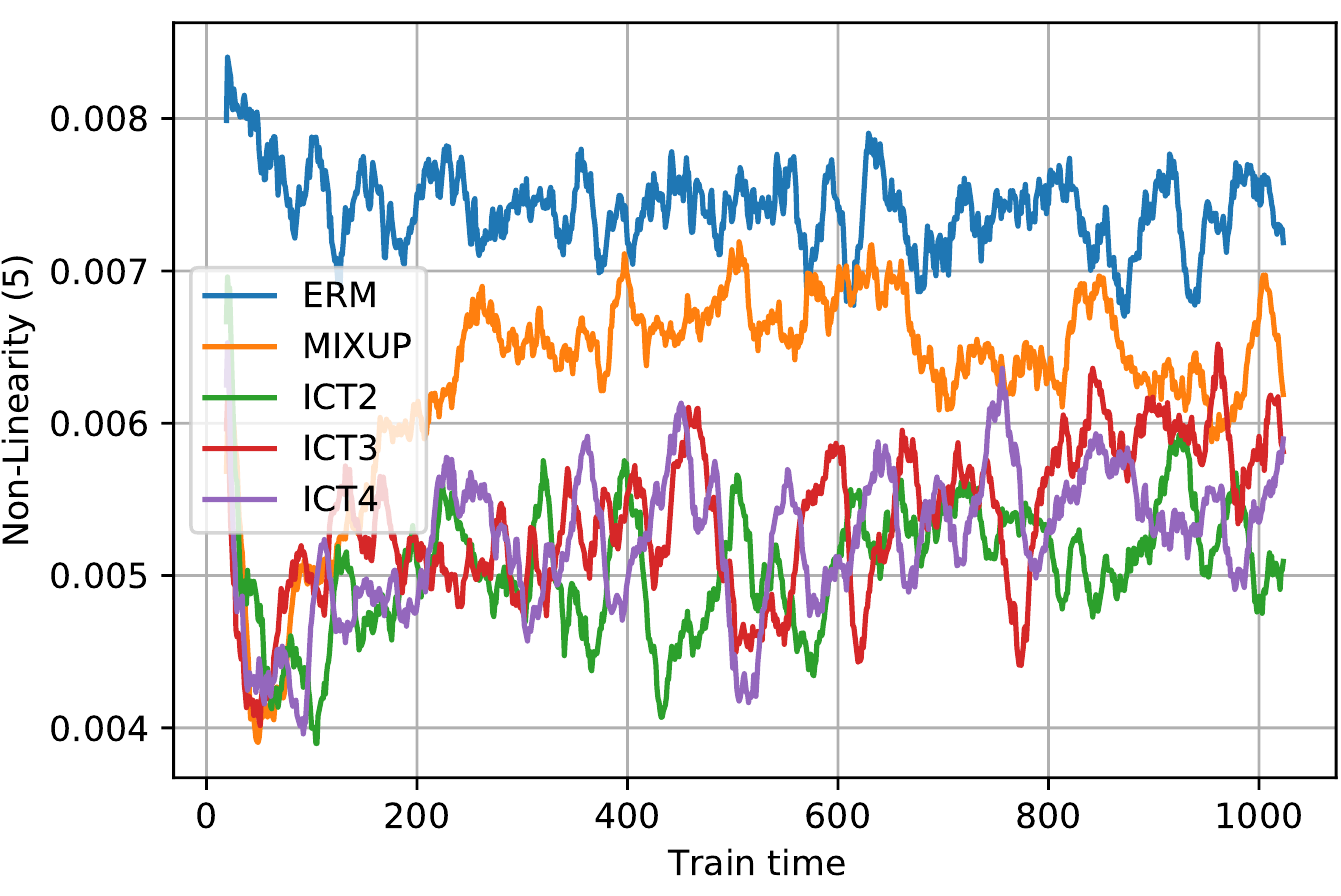}
      \caption{\small Layer 5}
\end{subfigure}

\negspace{-1mm}
\caption{\textbf{SVHN}: Linearity propagation in different layers with \LABELS\ labeled examples and consistency weight of \CTWEIGHT. ERM and mixup refer to the cases where only supervised loss is used. ICT[X] refer to a setup where the ICT loss with X mixup points are used as unsupervised loss. The supervised loss with the ICT does not use mixup.}
\label{fig:svhn_labels250_cw100_linearity}
\negspace{-3mm}
\end{figure*}

\begin{figure*}[t]
\def \LABELS {2000}
\def \CTWEIGHT {100}
\centering
\begin{subfigure}{.33\textwidth}
      \centering
      \includegraphics[width=.99\linewidth]{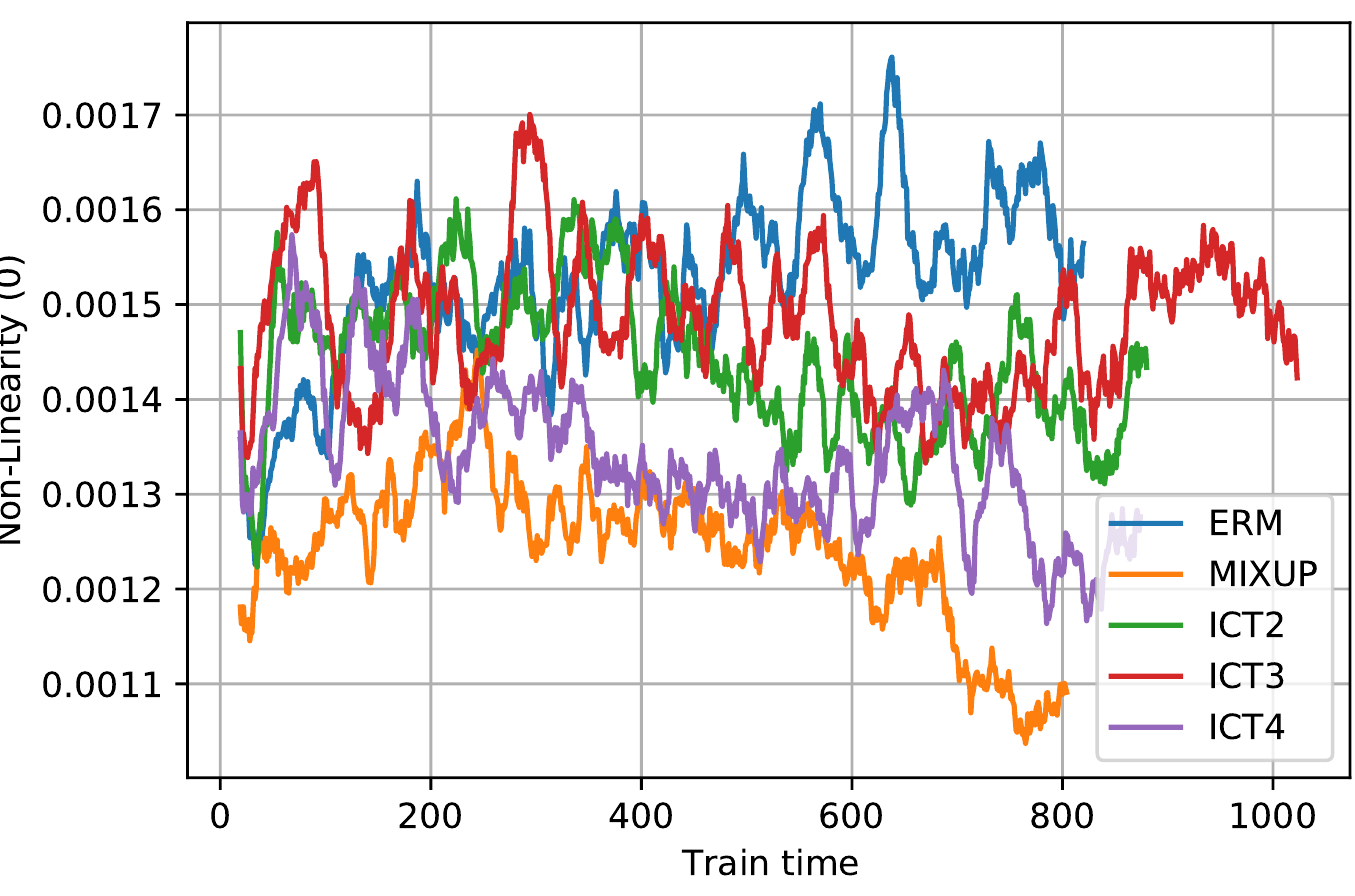}
      \caption{\small Layer 0}
\end{subfigure}\hfill
\begin{subfigure}{.33\textwidth}
      \centering
      \includegraphics[width=.99\linewidth]{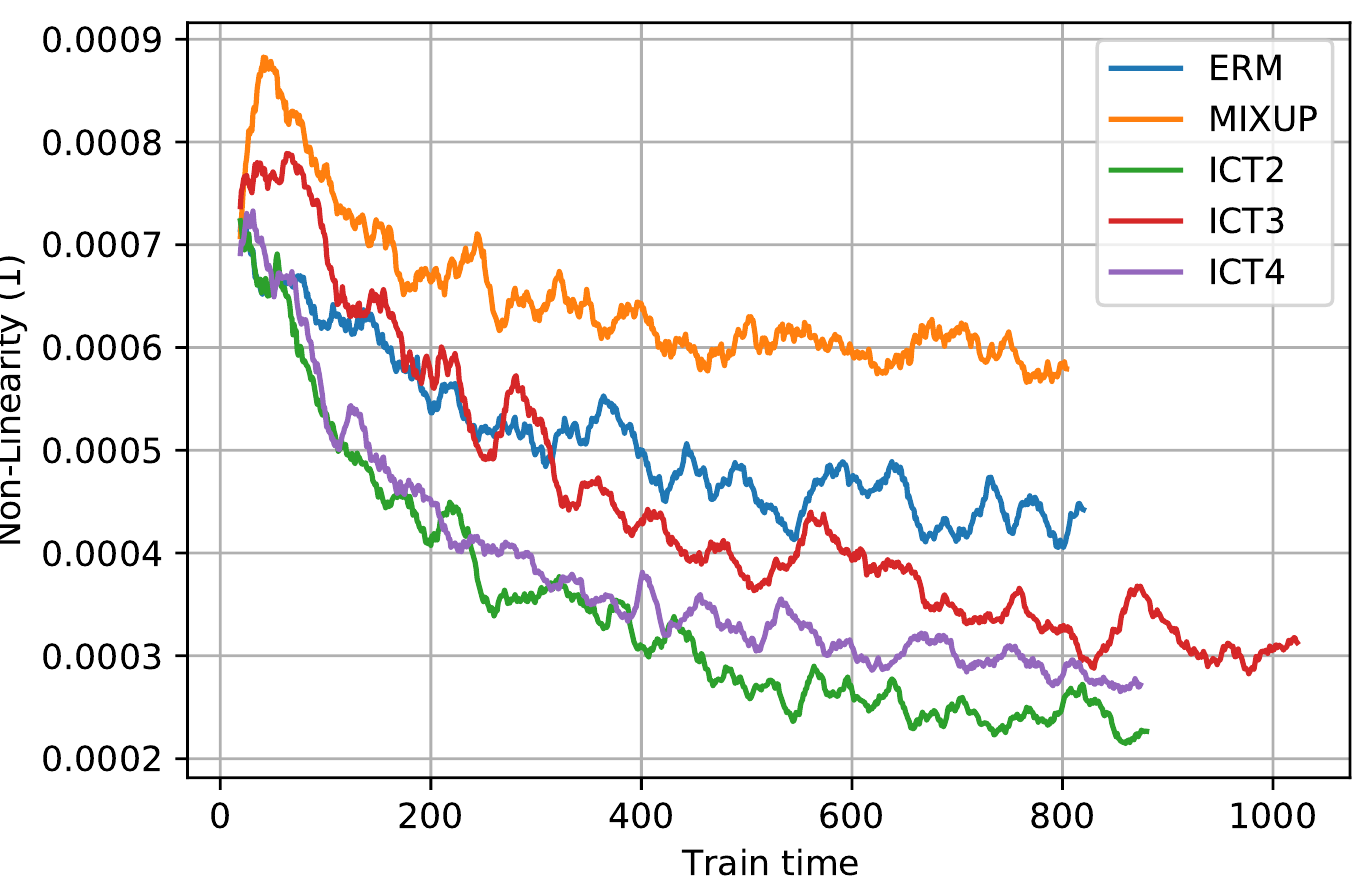}
      \caption{\small Layer 1}
\end{subfigure}\hfill
\begin{subfigure}{.33\textwidth}
      \centering
      \includegraphics[width=.99\linewidth]{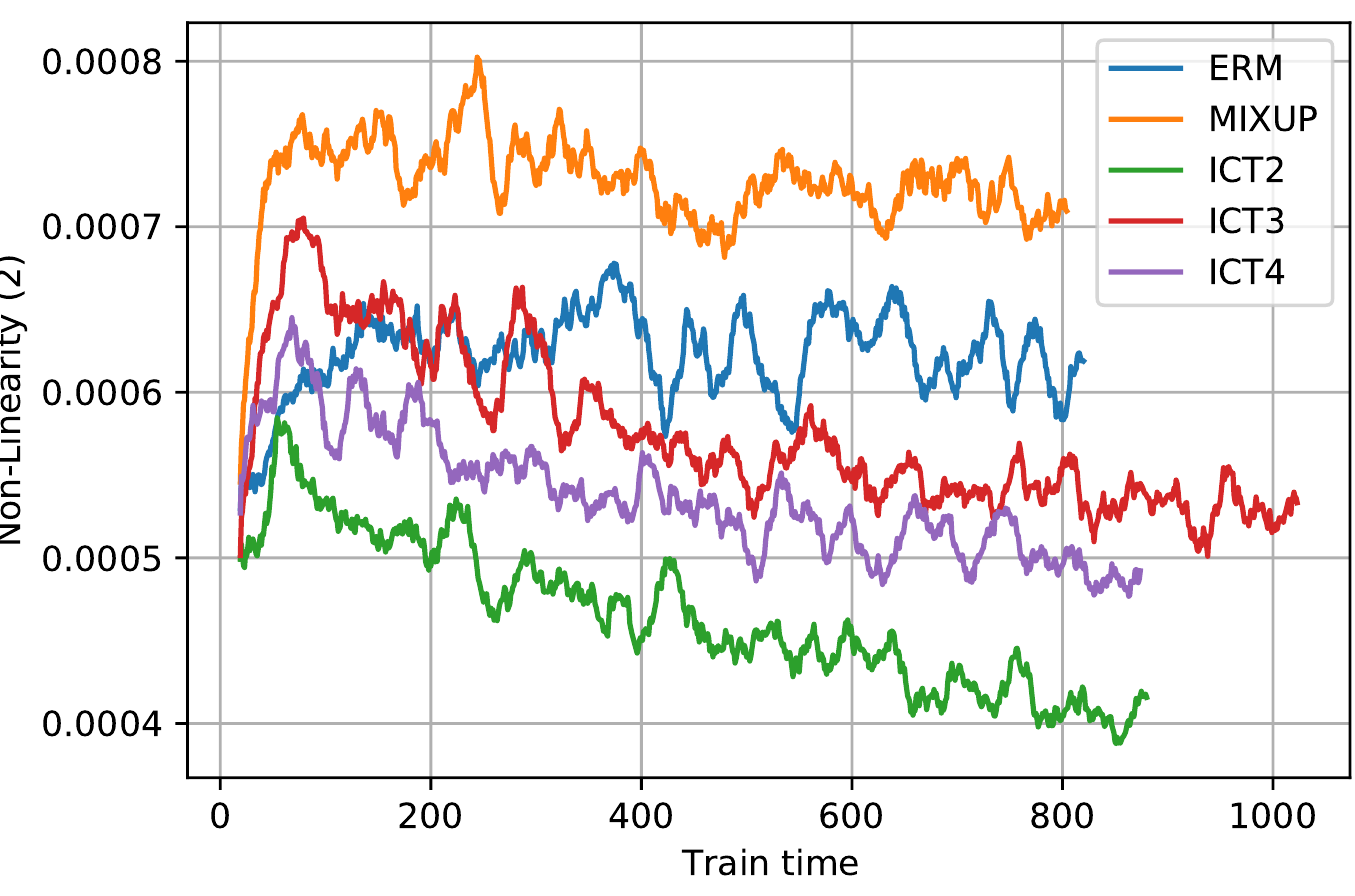}
      \caption{\small Layer 2}
\end{subfigure}

\begin{subfigure}{.33\textwidth}
      \centering
      \includegraphics[width=.99\linewidth]{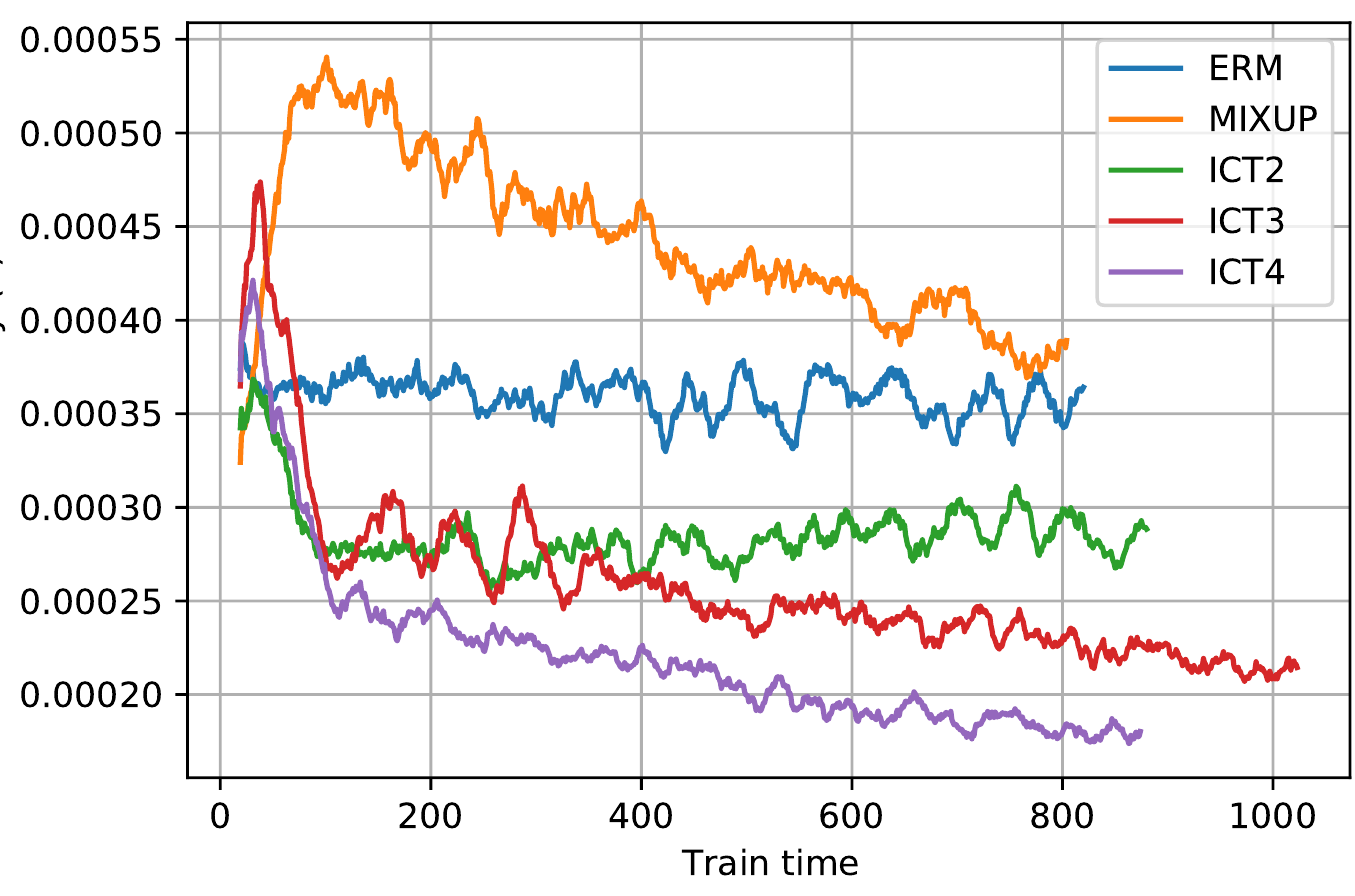}
      \caption{\small Layer 3}
\end{subfigure}\hfill
\begin{subfigure}{.33\textwidth}
      \centering
      \includegraphics[width=.99\linewidth]{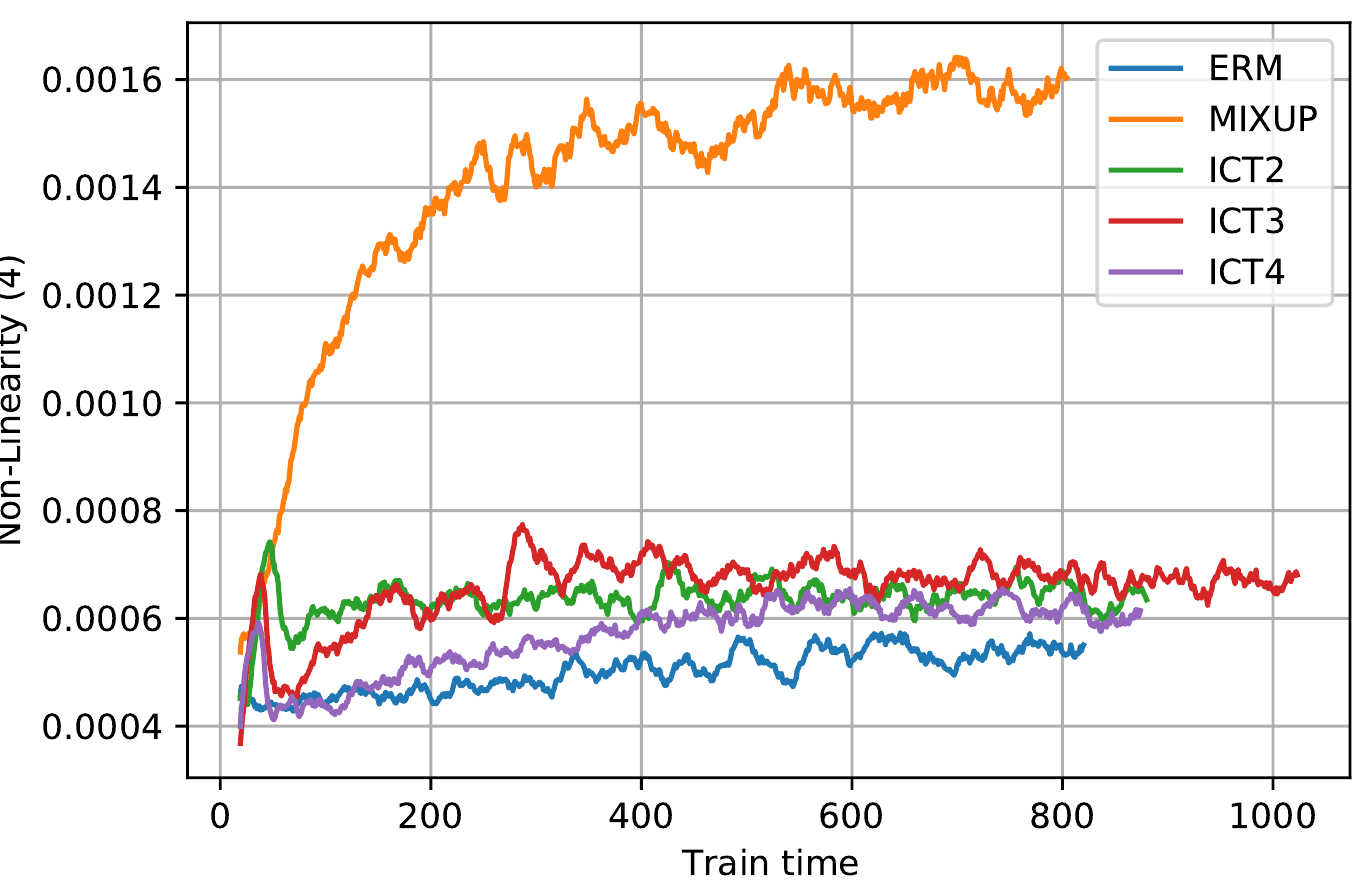}
      \caption{\small Layer 4}
\end{subfigure}\hfill
\begin{subfigure}{.33\textwidth}
      \centering
      \includegraphics[width=.99\linewidth]{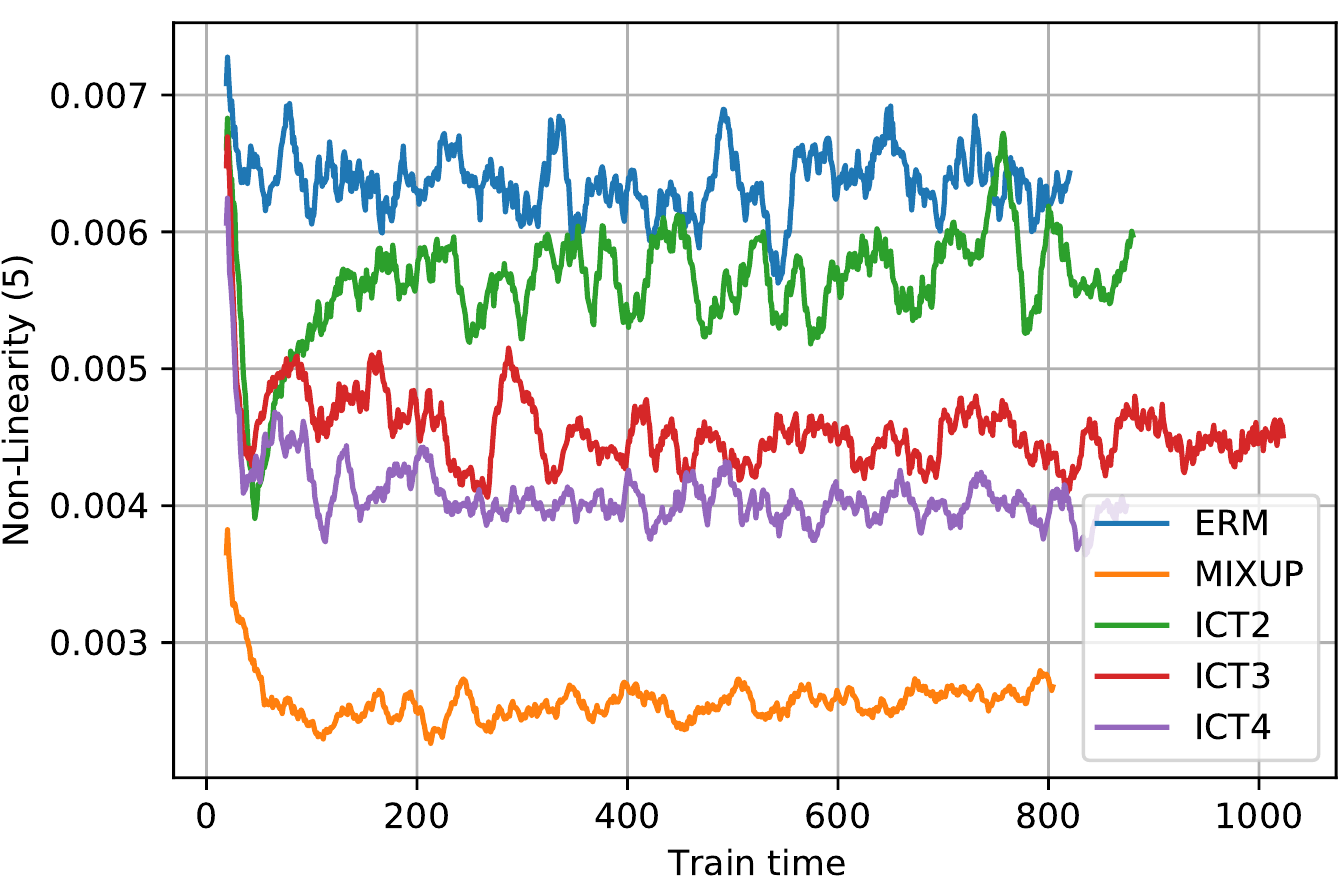}
      \caption{\small Layer 5}
\end{subfigure}

\negspace{-1mm}
\caption{\textbf{SVHN}: Linearity propagation in different layers with \LABELS\ labeled examples and consistency weight of \CTWEIGHT. ERM and mixup refer to the cases where only supervised loss is used. $\ictx_K$ refer to a setup where the ICT loss with $K$ mixup points are used as unsupervised loss. The supervised loss with the ICT does not use mixup.}
\label{fig:svhn_labels2000_cw100_linearity}
\negspace{-3mm}
\end{figure*}

\begin{figure*}[t]
\def \LABELS {4000}
\def \CTWEIGHT {100}
\centering
\begin{subfigure}{.33\textwidth}
      \centering
      \includegraphics[width=.99\linewidth]{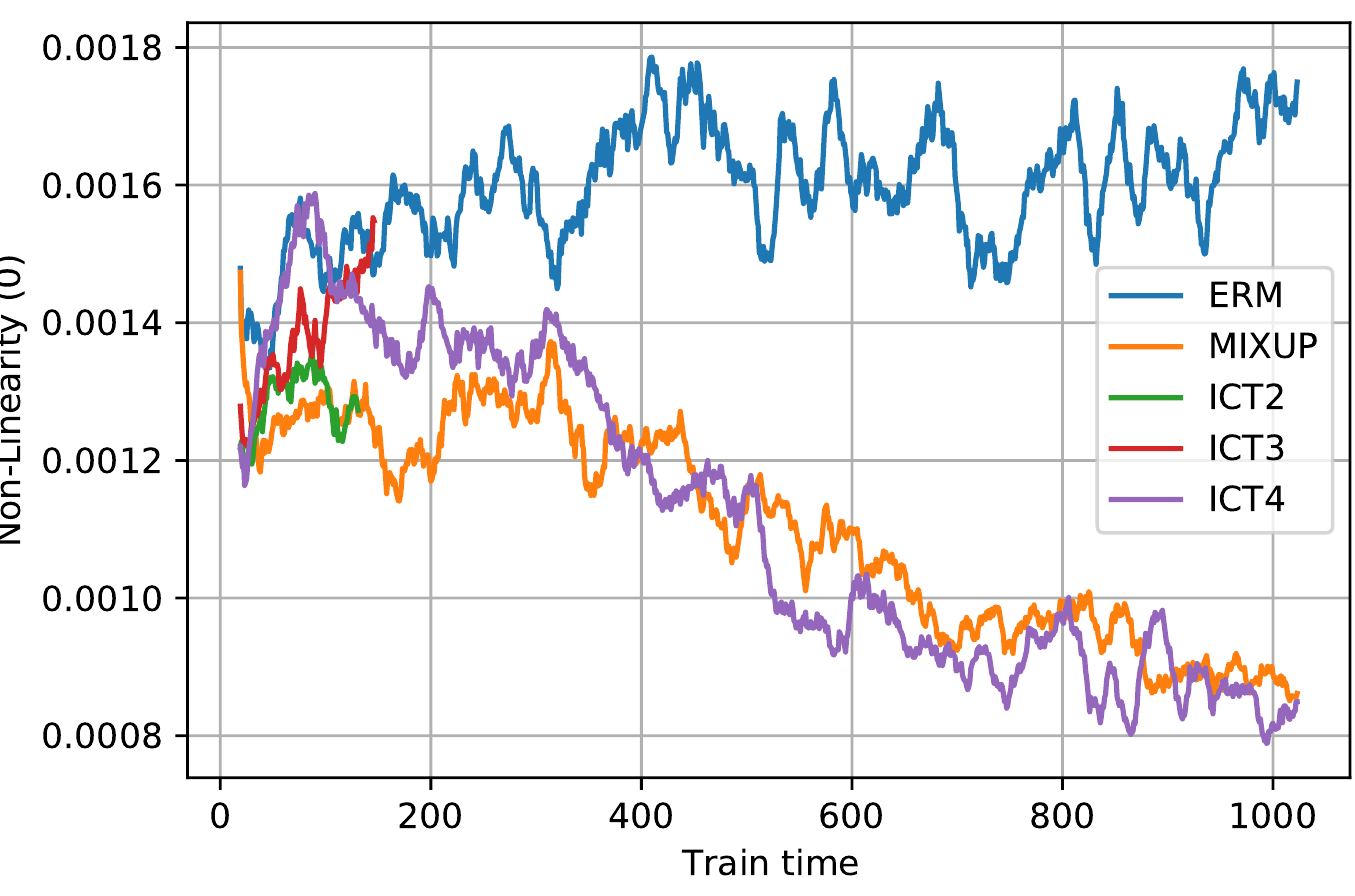}
      \caption{\small Layer 0}
\end{subfigure}\hfill
\begin{subfigure}{.33\textwidth}
      \centering
      \includegraphics[width=.99\linewidth]{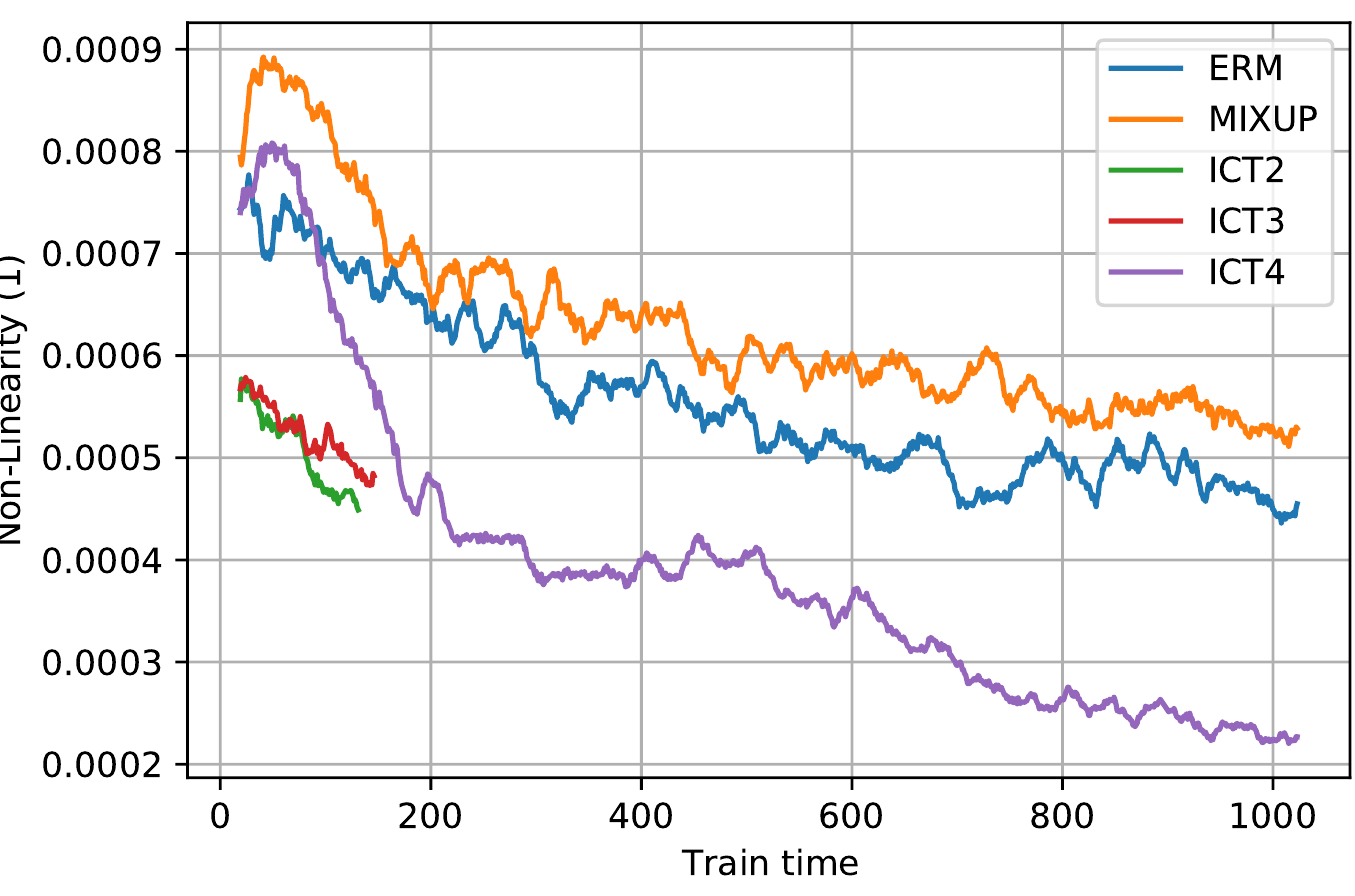}
      \caption{\small Layer 1}
\end{subfigure}\hfill
\begin{subfigure}{.33\textwidth}
      \centering
      \includegraphics[width=.99\linewidth]{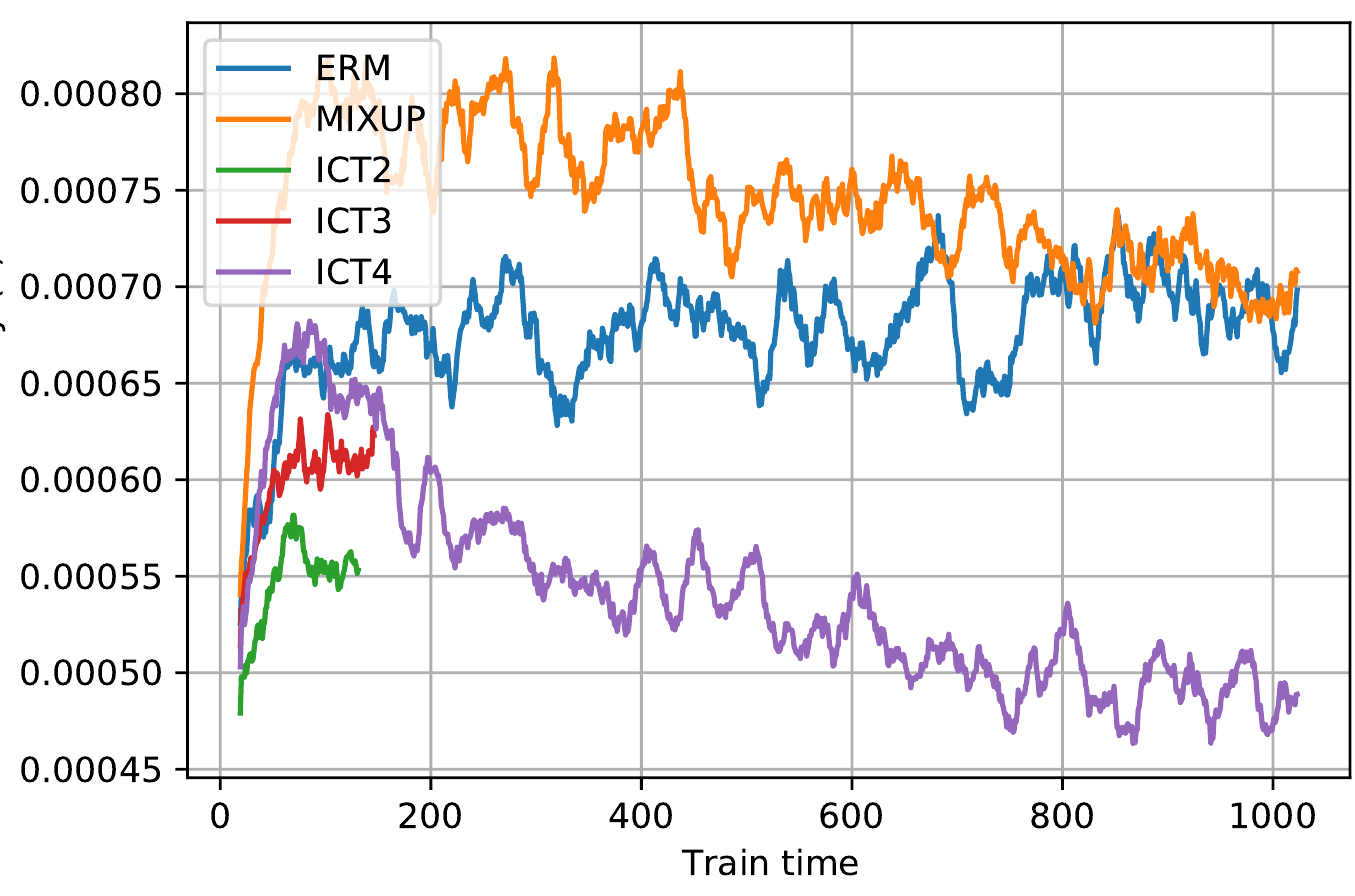}
      \caption{\small Layer 2}
\end{subfigure}

\begin{subfigure}{.33\textwidth}
      \centering
      \includegraphics[width=.99\linewidth]{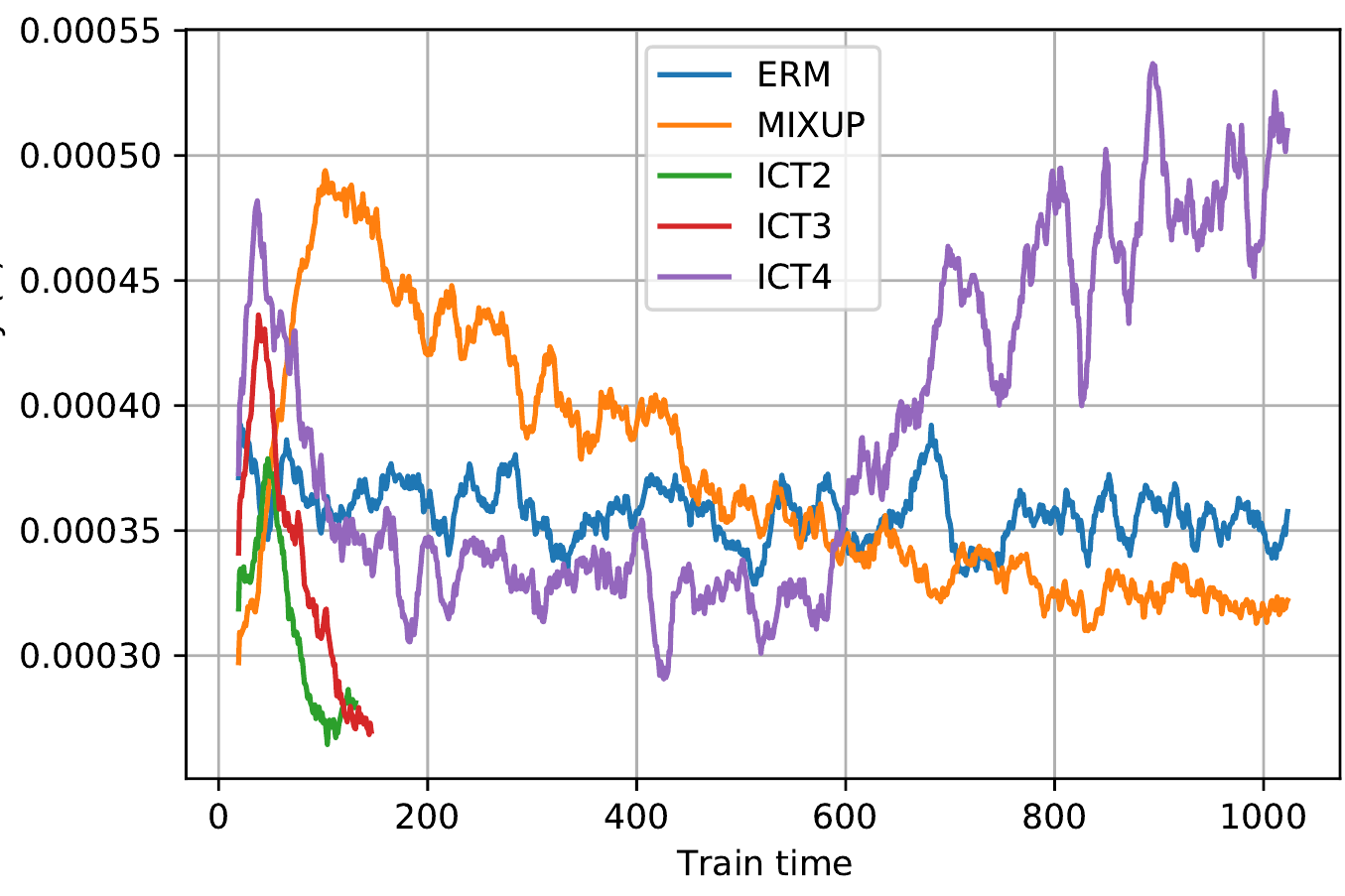}
      \caption{\small Layer 3}
\end{subfigure}\hfill
\begin{subfigure}{.33\textwidth}
      \centering
      \includegraphics[width=.99\linewidth]{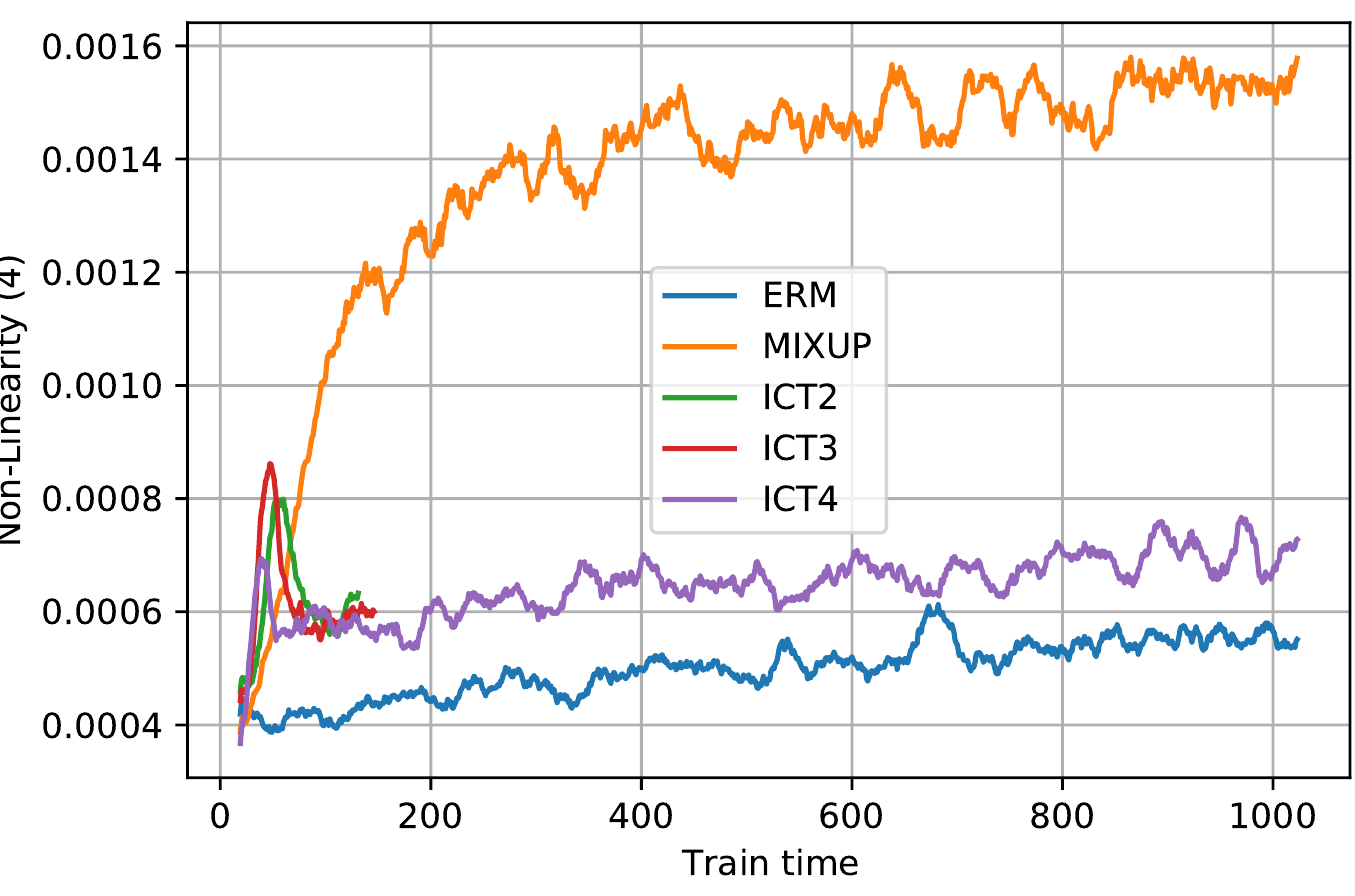}
      \caption{\small Layer 4}
\end{subfigure}\hfill
\begin{subfigure}{.33\textwidth}
      \centering
      \includegraphics[width=.99\linewidth]{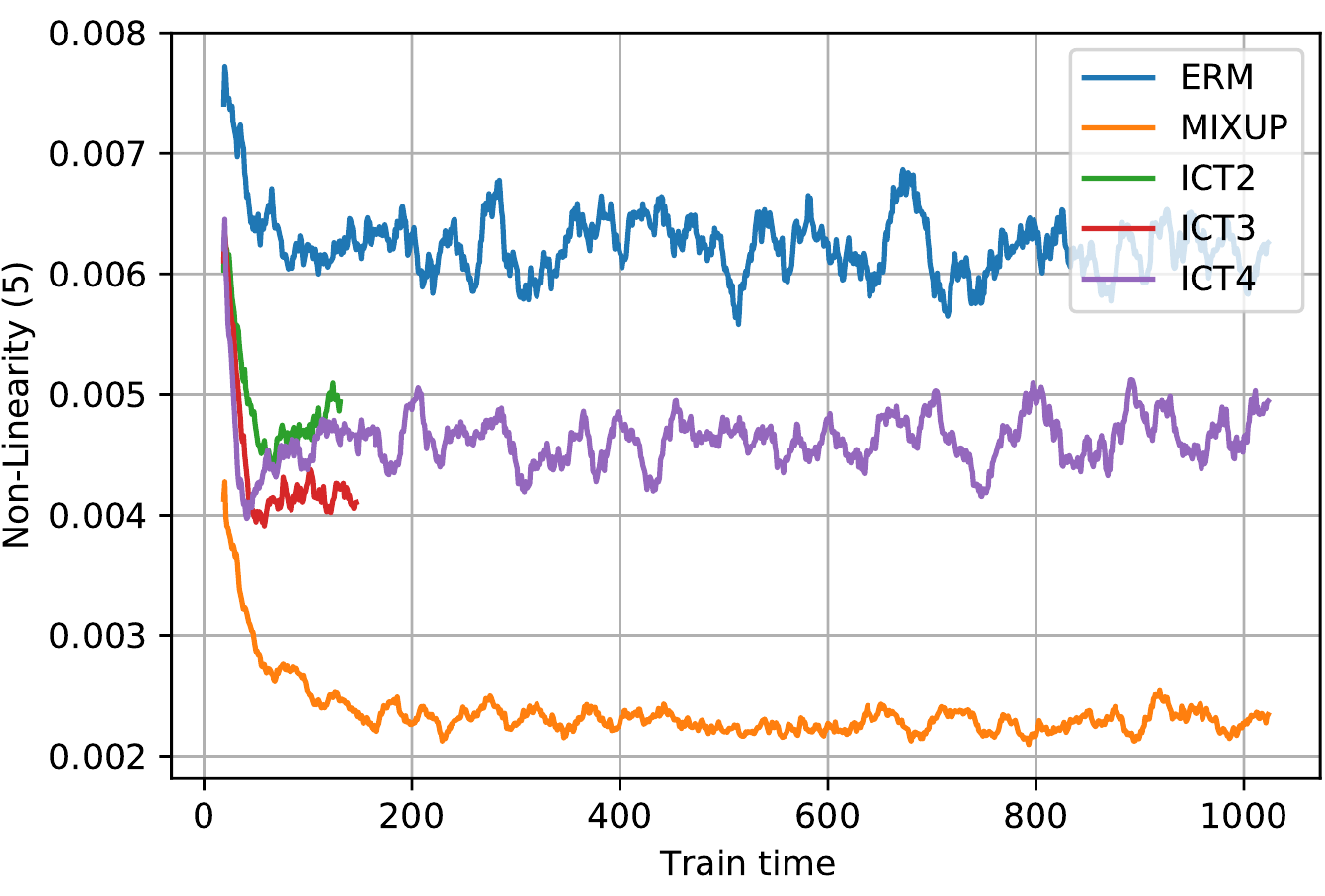}
      \caption{\small Layer 5}
\end{subfigure}

\negspace{-1mm}
\caption{\textbf{SVHN}: Linearity propagation in different layers with \LABELS\ labeled examples and consistency weight of \CTWEIGHT. ERM and mixup refer to the cases where only supervised loss is used. $\ictx_K$ refer to a setup where the ICT loss with $K$ mixup points are used as unsupervised loss. The supervised loss with the ICT does not use mixup.}
\label{fig:svhn_labels4000_cw100_linearity}
\negspace{-3mm}
\end{figure*}

\begin{figure*}[t]
\def \LABELS {100}
\def \CTWEIGHT {100}
\centering
\begin{subfigure}{.33\textwidth}
      \centering
      \includegraphics[width=.99\linewidth]{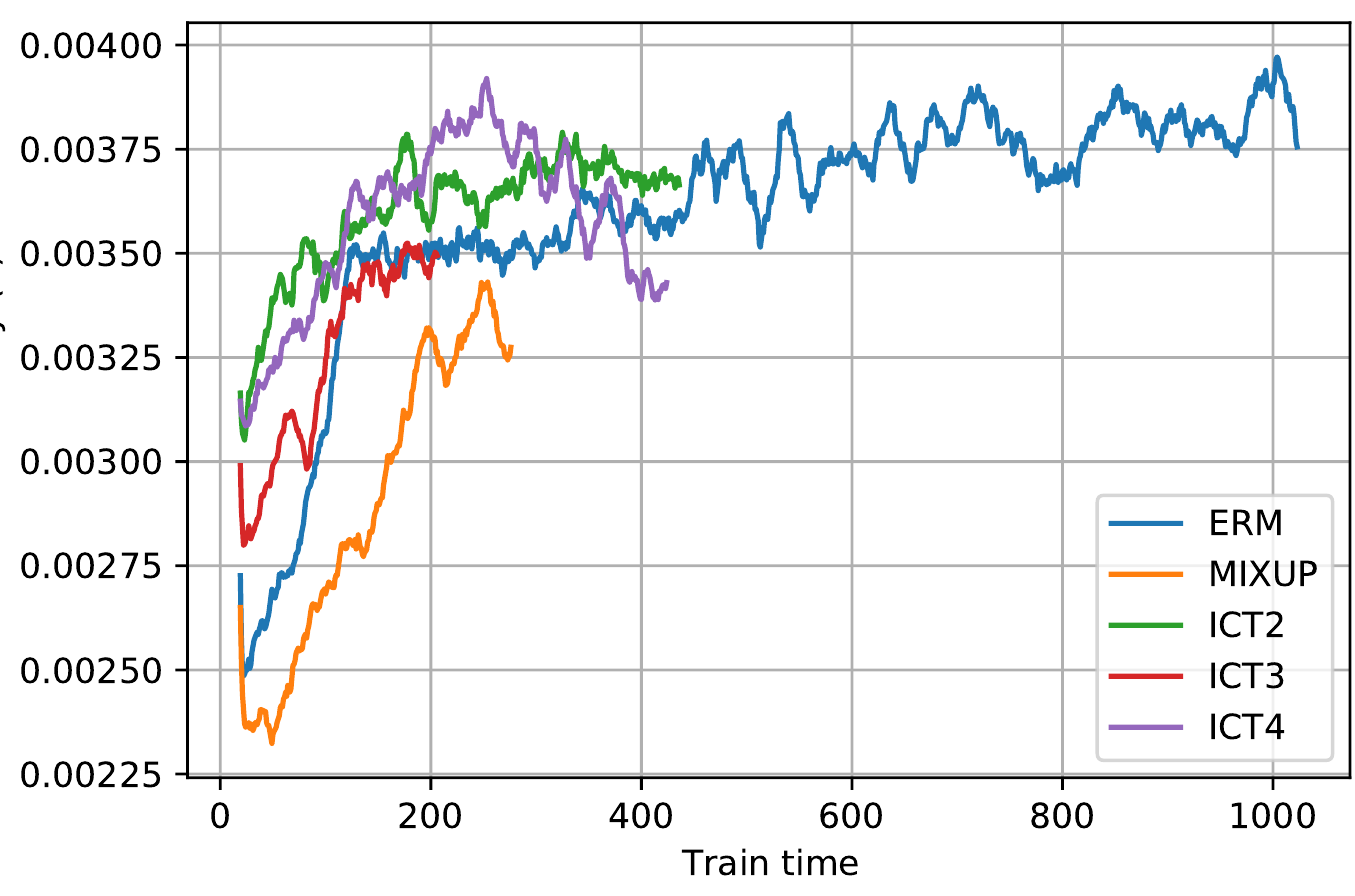}
      \caption{\small Layer 0}
\end{subfigure}\hfill
\begin{subfigure}{.33\textwidth}
      \centering
      \includegraphics[width=.99\linewidth]{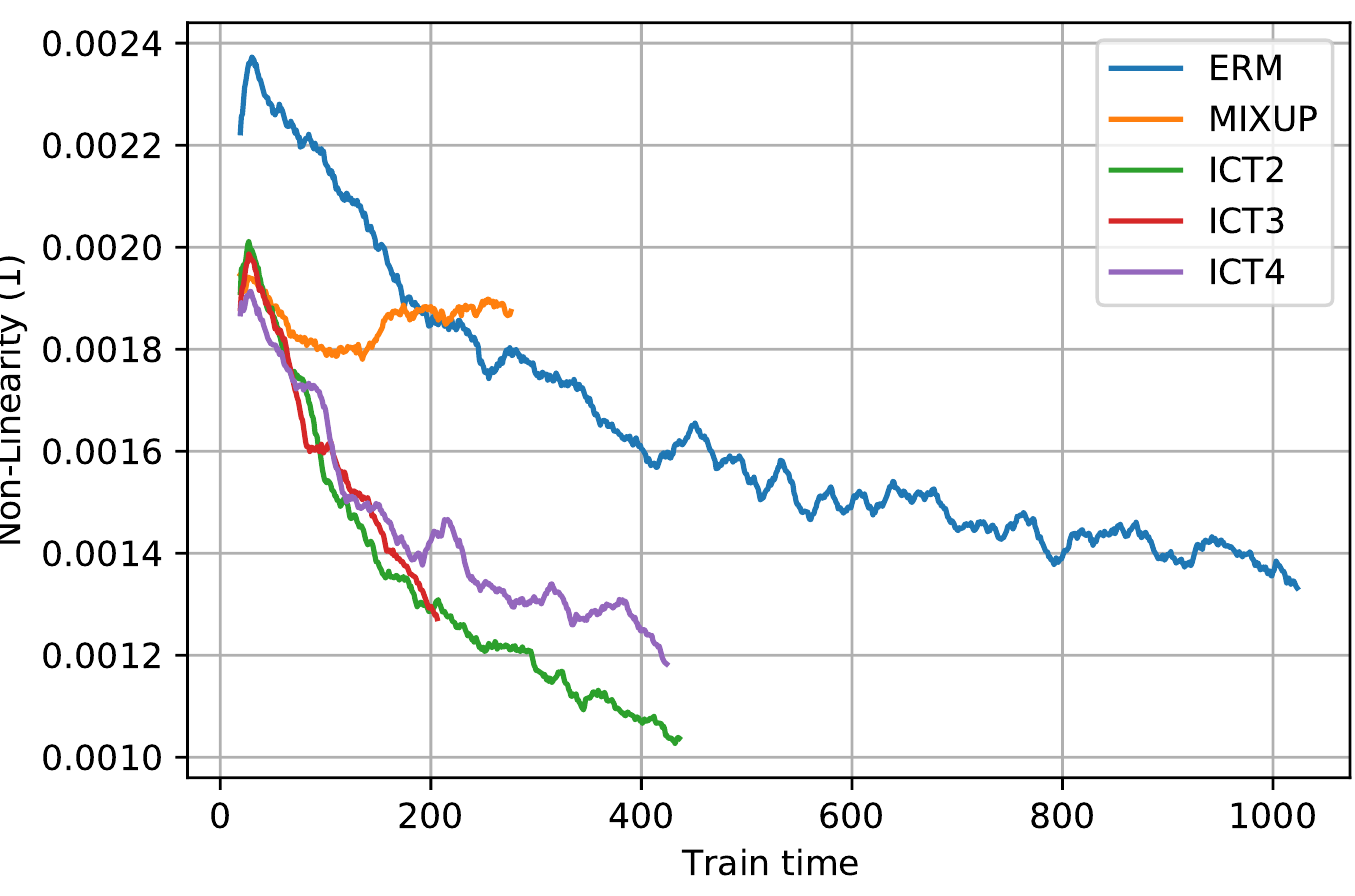}
      \caption{\small Layer 1}
\end{subfigure}\hfill
\begin{subfigure}{.33\textwidth}
      \centering
      \includegraphics[width=.99\linewidth]{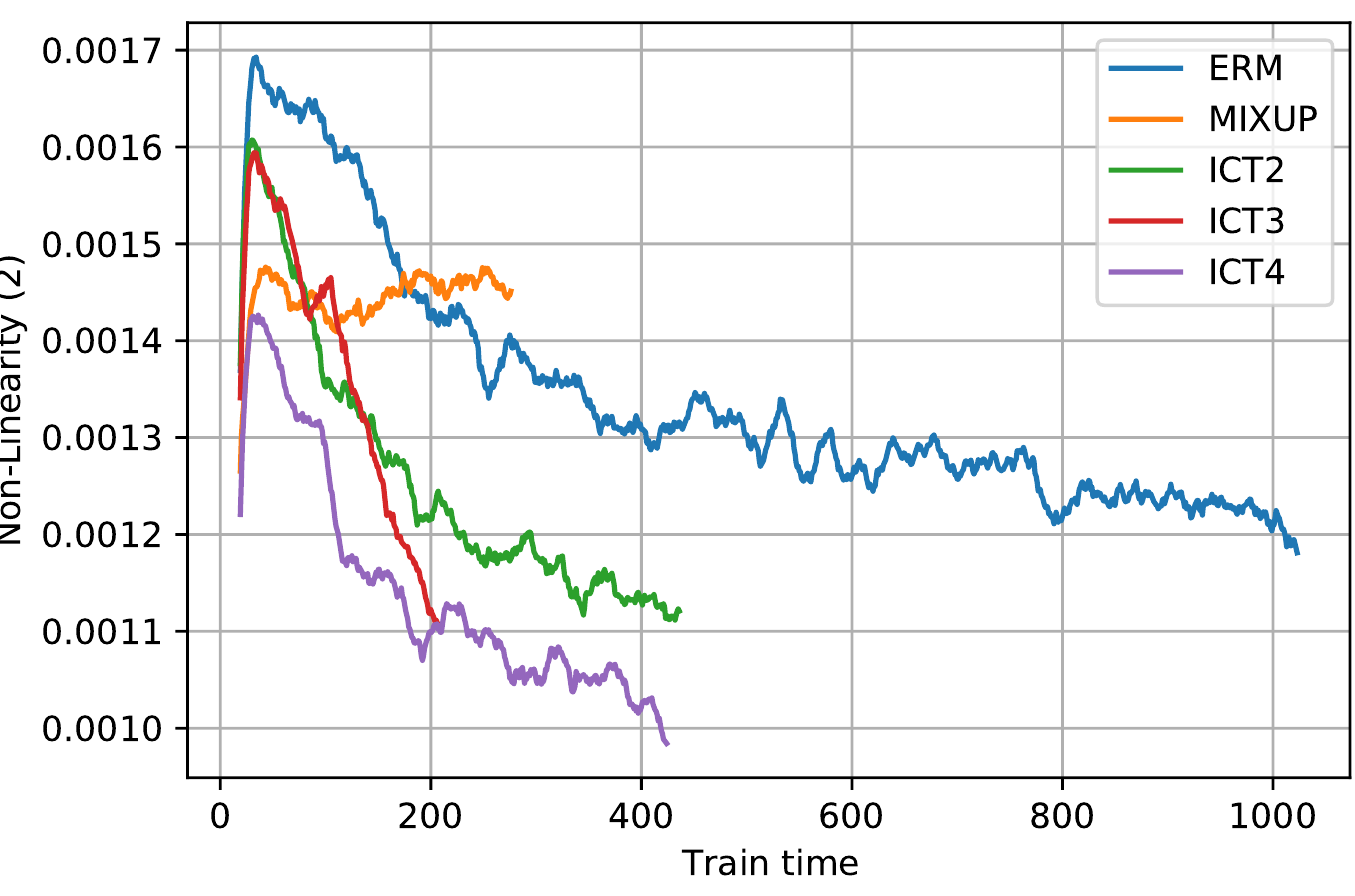}
      \caption{\small Layer 2}
\end{subfigure}

\begin{subfigure}{.33\textwidth}
      \centering
      \includegraphics[width=.99\linewidth]{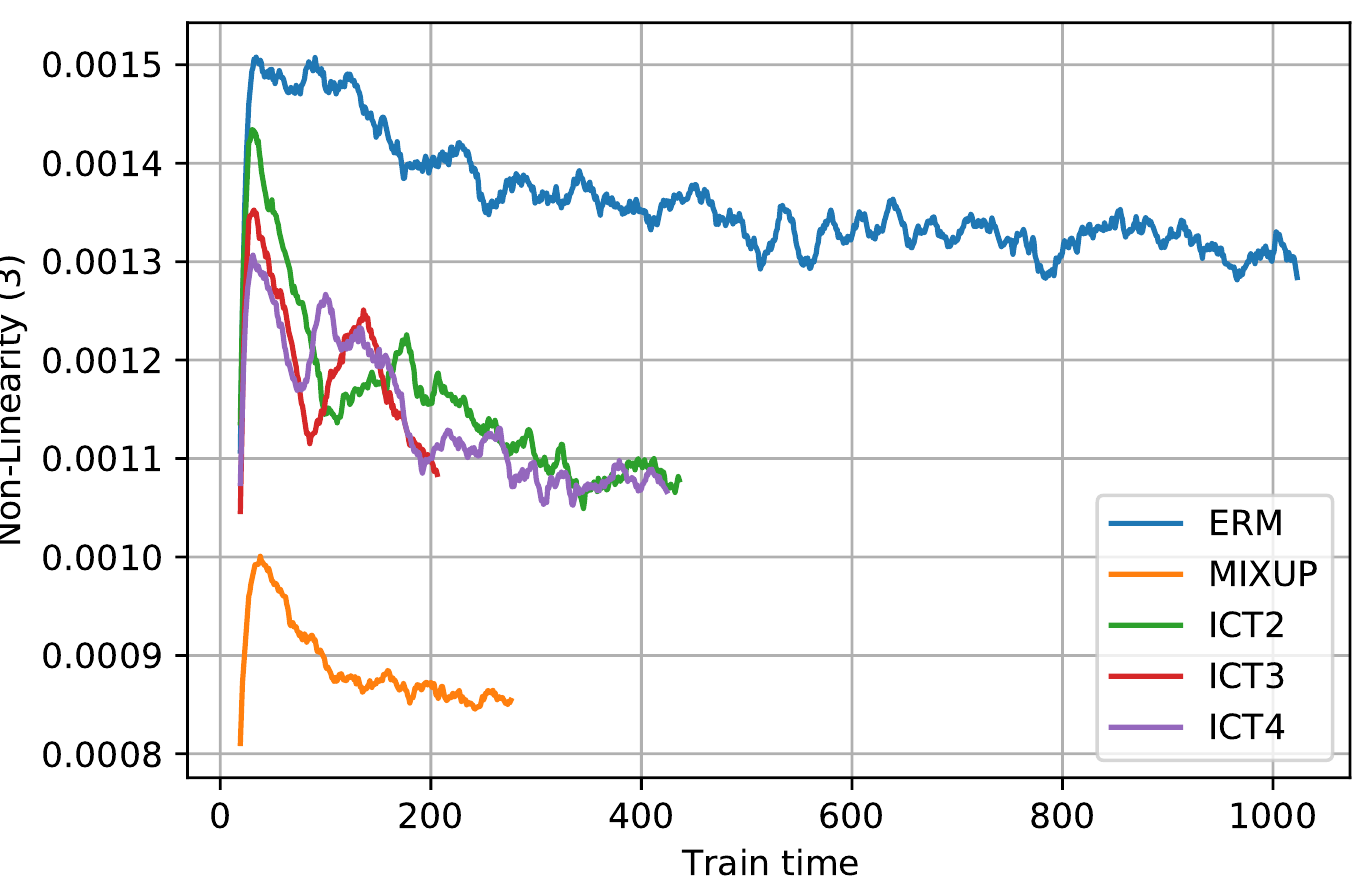}
      \caption{\small Layer 3}
\end{subfigure}\hfill
\begin{subfigure}{.33\textwidth}
      \centering
      \includegraphics[width=.99\linewidth]{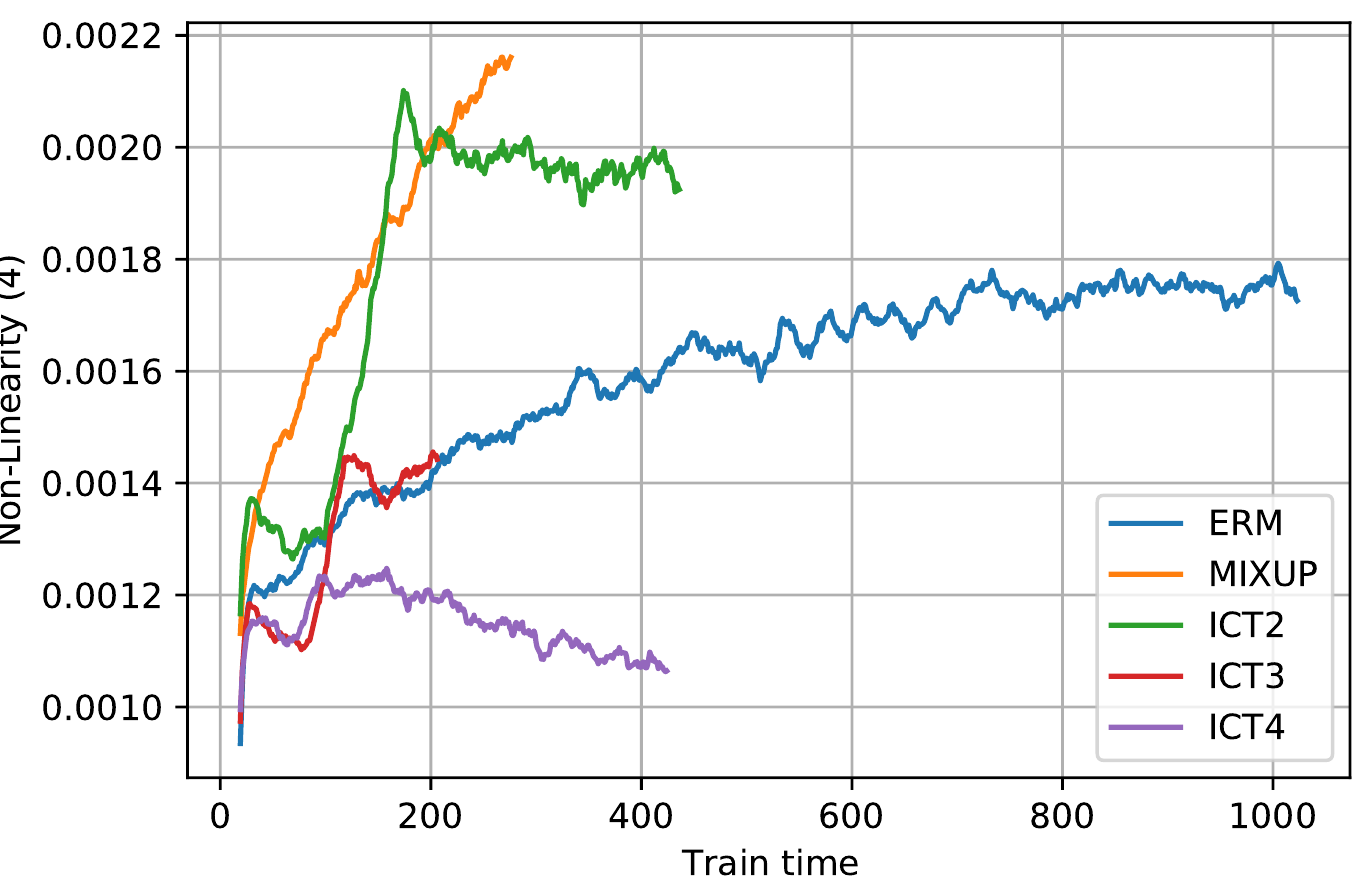}
      \caption{\small Layer 4}
\end{subfigure}\hfill
\begin{subfigure}{.33\textwidth}
      \centering
      \includegraphics[width=.99\linewidth]{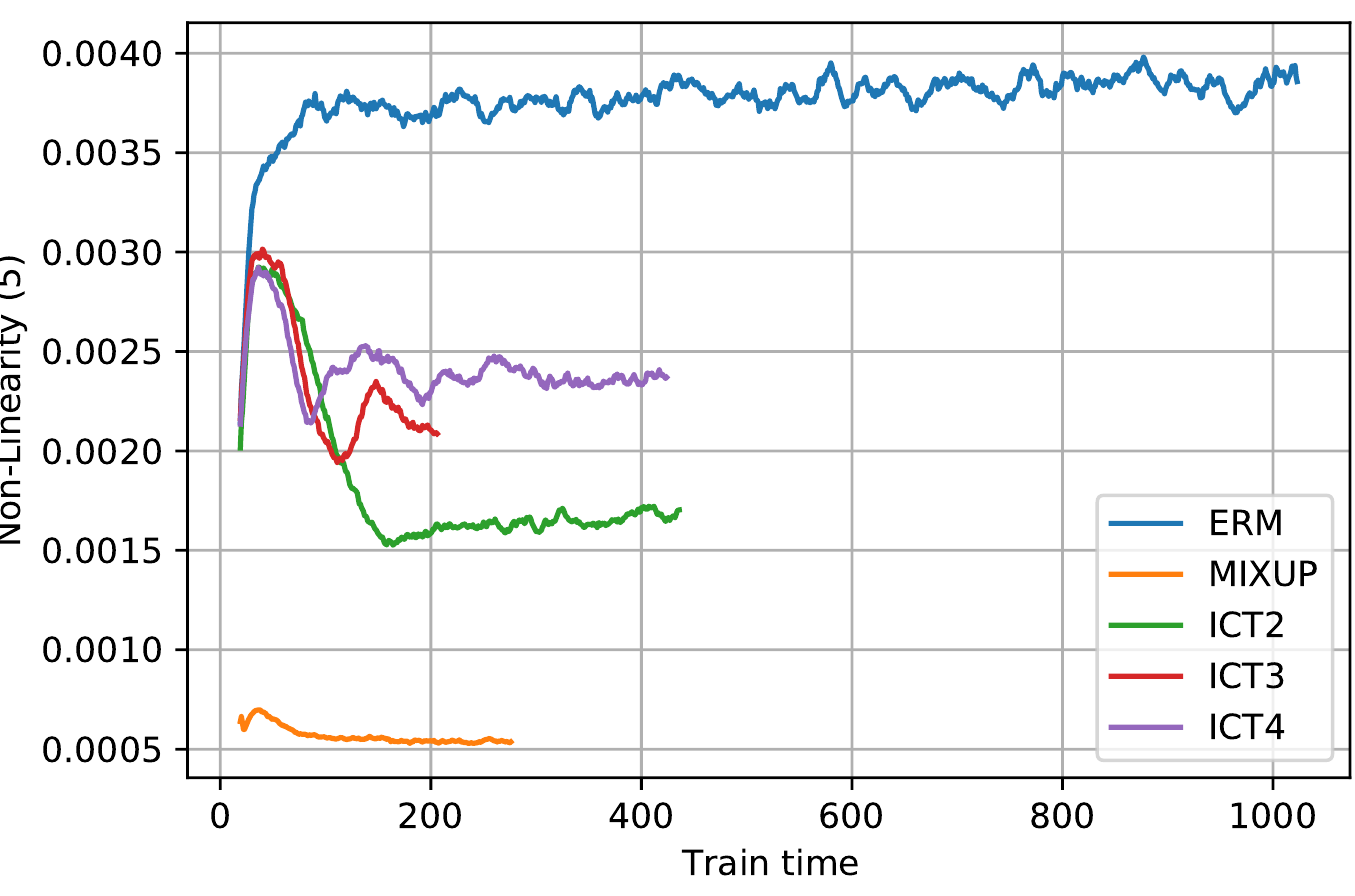}
      \caption{\small Layer 5}
\end{subfigure}

\negspace{-1mm}
\caption{\textbf{CIFAR-100}: Linearity propagation in different layers with \LABELS\ labeled examples and consistency weight of \CTWEIGHT. ERM and mixup refer to the cases where only supervised loss is used. $\ictx_K$ refer to a setup where the ICT loss with $K$ mixup points are used as unsupervised loss. The supervised loss with the ICT does not use mixup.}
\label{fig:cifar100_labels1000_cw100_linearity}
\negspace{-3mm}
\end{figure*}

\begin{figure*}[t]
\centering
\includegraphics[scale=0.7]{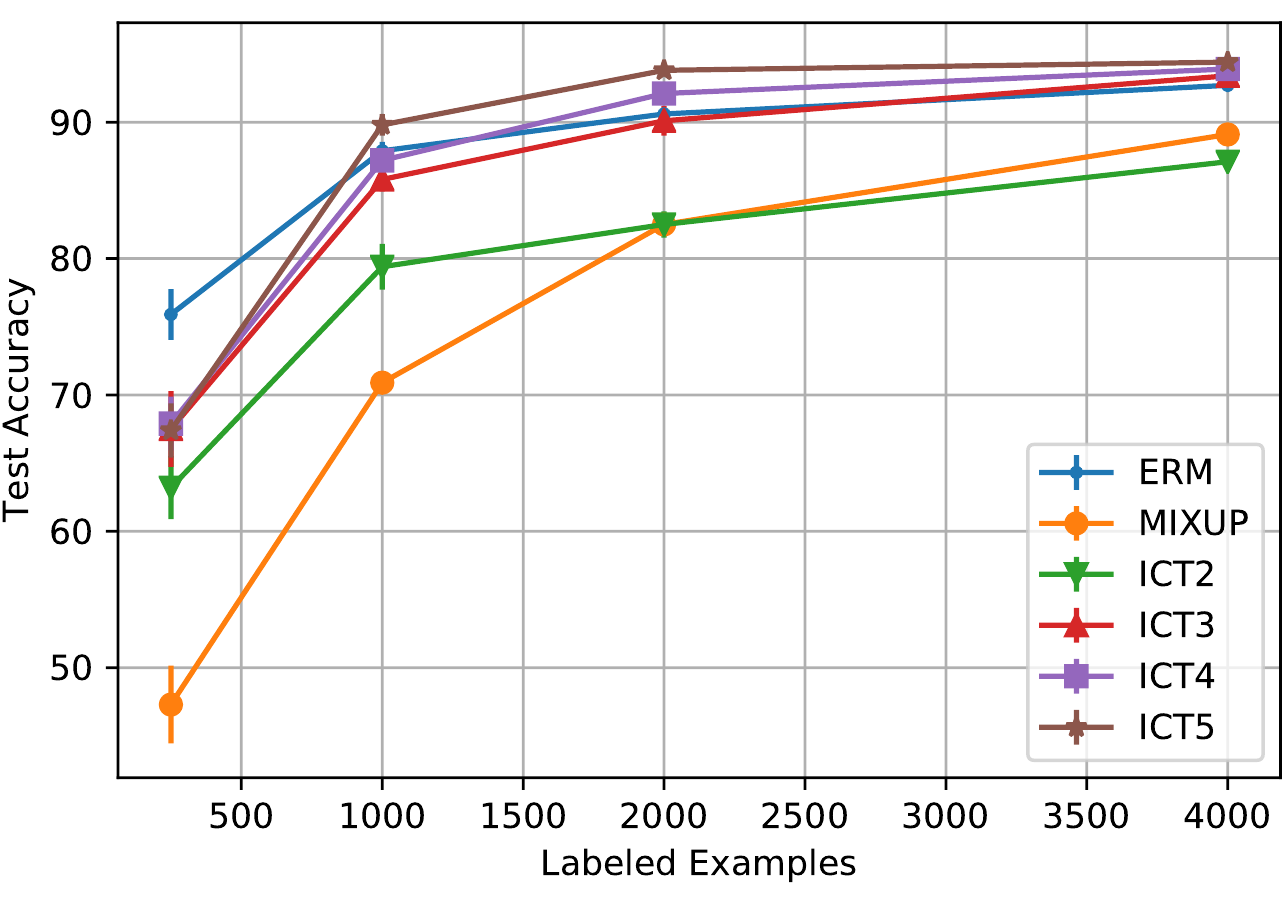}
\negspace{-1mm}
\caption{\textbf{SVHN}: Test accuracy when stronger linearity is enforced. ERM and mixup refer to the cases where only supervised loss is used. ICT[X] refer to a setup where the ICT loss with X mixup points are used as unsupervised loss. The supervised loss with the ICT does not use mixup loss.}
\label{fig:svhn_ictX}
\negspace{-3mm}
\end{figure*}

\begin{figure*}[t]
\centering
\includegraphics[scale=0.7]{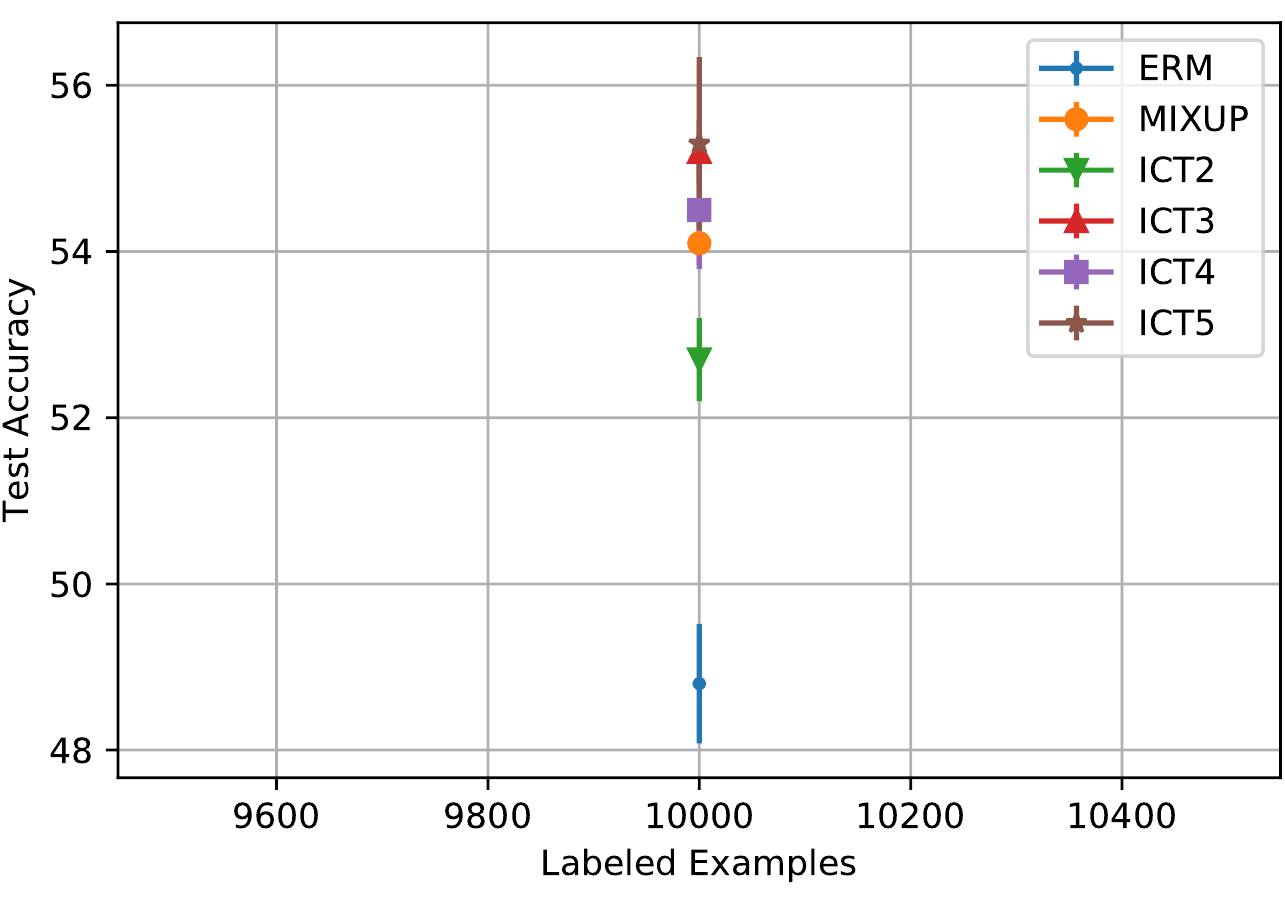}
\negspace{-1mm}
\caption{\textbf{CIFAR-100}: Test accuracy}
\label{fig:cifar100_ictX}
\negspace{-3mm}
\end{figure*}

\end{document}